\documentclass[11pt,letterpaper]{article}

\usepackage[top=1in,bottom=1in,left=1in,right=1in]{geometry}
\usepackage{natbib}
\setcitestyle{numbers,square,comma}

\usepackage[normalem]{ulem}

\usepackage[utf8]{inputenc}
\usepackage[T1]{fontenc}
\usepackage{hyperref}
\usepackage{url}
\usepackage{booktabs}
\usepackage{amsfonts}
\usepackage{nicefrac}
\usepackage{microtype}
\usepackage{xcolor}

\usepackage{amsmath,amssymb}
\usepackage{graphicx}
\graphicspath{{figs/}}
\usepackage{subcaption}
\usepackage{multirow}
\usepackage{mathtools}
\usepackage{amsthm}
\usepackage[capitalize,noabbrev]{cleveref}
\usepackage{placeins}
\usepackage{wrapfig}
\usepackage{enumitem}

\theoremstyle{plain}
\newtheorem{theorem}{Theorem}[section]

\theoremstyle{definition}

\theoremstyle{definition}

\usepackage[most]{tcolorbox}
\usepackage{aliascnt}
\newaliascnt{proposition}{theorem}
\newtheorem{proposition}[proposition]{Proposition}
\aliascntresetthe{proposition}
\crefname{proposition}{Proposition}{Propositions}
\Crefname{proposition}{Proposition}{Propositions}
\theoremstyle{plain}
\newaliascnt{lemma}{theorem}
\newtheorem{lemma}[lemma]{Lemma}
\aliascntresetthe{lemma}
\crefname{lemma}{Lemma}{Lemmas}
\Crefname{lemma}{Lemma}{Lemmas}
\theoremstyle{remark}
\newaliascnt{remark}{theorem}
\newtheorem{remark}[remark]{Remark}
\aliascntresetthe{remark}
\crefname{remark}{Remark}{Remarks}
\Crefname{remark}{Remark}{Remarks}
\newtheoremstyle{boldremarklabel}
  {0.5\topsep}{0.5\topsep}{\normalfont}{}{\itshape}{.}{0.5em}
  {\textnormal{\bfseries\thmname{#1}\thmnumber{ #2}}\textnormal{\thmnote{ (#3)}}}
\theoremstyle{boldremarklabel}
\newtheorem{emphasizedremark}[remark]{Remark}
\theoremstyle{definition}
\newcounter{phenomenon}

\newtcolorbox{phenomenon}[1]{
    enhanced,
    breakable,
    colback=white,
    colframe=black!78,
    boxrule=0.7pt,
    arc=5pt,
    outer arc=5pt,
    left=8pt,
    right=8pt,
    top=6pt,
    bottom=6pt,
    before skip=0.9em,
    after skip=0.9em,
    before upper={%
        \refstepcounter{phenomenon}%
        {\bfseries #1}\par
        \vspace{2.5pt}%
        {\color{black!55}\hrule height 0.4pt}%
        \vspace{4.5pt}%
    }
    }

\title{Extrapolation across Phase Transitions with Polynomial and Kernel Regression: Error Bounds and Power-Law Scaling}

\author{%
\begin{tabular}{@{}ccc@{}}
Jeffrey Wei & Manolis Zampetakis & John Sous \\[6pt]
Yale University & Yale University & Yale University \\[3pt]
\small\href{mailto:jeffrey.wei@yale.edu}{\texttt{jeffrey.wei@yale.edu}} &
\small\href{mailto:manolis.zampetakis@yale.edu}{\texttt{manolis.zampetakis@yale.edu}} &
\small\href{mailto:john.sous@yale.edu}{\texttt{john.sous@yale.edu}}
\end{tabular}
}
\date{}

\makeatletter
\let\tocentry\addcontentsline
\renewcommand{\addcontentsline}[3]{}
\makeatother
\hypersetup{
  colorlinks=true,
  linkcolor=red!80!black,
  citecolor=blue,
  urlcolor=magenta,
  pdftitle={Extrapolation across Phase Transitions with Polynomial and Kernel Regression: Error Bounds and Power-Law Scaling},
  pdfauthor={Jeffrey Wei, Manolis Zampetakis, John Sous}
}
\allowdisplaybreaks
\begin{document}
\maketitle
\begin{abstract}
Phase transitions occur when a physical system undergoes a transformation as it crosses a critical point. We study whether regression methods trained only in one phase can predict the physical properties of an unobserved phase. Across analytical, numerically simulated, and experimental transitions, we find that polynomial and kernel regression recover the unobserved phase structure, while neural networks fail. We quantify extrapolation reliability through the transfer coefficient, defined as the ratio of the mean squared error (MSE) on the extrapolation region to that on the training region. We use the generalized eigenvalue problem (GEVP) to derive an upper bound on the worst-case transfer coefficient without requiring labels from the extrapolation region. Empirically, the GEVP bound tightly upper bounds the empirical transfer coefficient across training-set sizes and extrapolation distance. We show that both the empirical transfer coefficient and GEVP bound grow as a power law in the extrapolation distance.
See \href{https://jwei302.github.io/extrapolation/}{project webpage} and \href{https://github.com/jwei302/extrapolation}{code}.
\end{abstract}

\section{Introduction}
\label{sec:intro}

A major goal of theoretical physics is to make predictions in regimes where information is limited or inaccessible. This objective can be reframed as an extrapolation problem: can we infer behavior in regimes where the physics is intractable from regimes where it is tractable?  Continuous phase transitions are a canonical example, where a smooth physical observable gives rise to nonanalytic behavior in derived quantities, in particular the order parameter, at the transition point~\citep{goldenfeld1992lectures}. Understanding phase transitions is important across science, spanning the behavior of matter in the early universe, the properties of magnets and superconductors, and phase structure in learning theory.

\citet{vargas2018extrapolating} and subsequent work~\citep{vargas2020physical,dai2023neural,kairon2024extrapolation} empirically showed that Gaussian processes can extrapolate across phase transitions. However, their approach evaluates extrapolation against target-phase ground truth rather than bounding the error before the data is available. This limits its usefulness in settings where the target phase is experimentally inaccessible or theoretically unresolved. 

We develop bounds for extrapolation across phase transitions without any labels from the target phase. We train polynomial regression, Gaussian processes (GPs), random Fourier features (RFFs), and neural network regression on data from an observed phase and ask whether they can extrapolate the physical observable into an unobserved phase. We evaluate this question across five continuous transitions, spanning analytical, numerically simulated, and noisy experimental data.  Across all transitions, we find that polynomial regression, RFFs, and GPs extrapolate the physical observable and the derived nonanalytic quantity across the phase boundary, while neural networks do not. 

We measure extrapolation reliability with the transfer coefficient, defined as the ratio of the mean squared error (MSE) on the extrapolation region to the training region. For finite feature models, we upper bound the transfer coefficient by solving a generalized eigenvalue problem (GEVP) on the feature covariance between the training and extrapolation region. Empirically, our GEVP bound remains nearly constant relative to the observed transfer coefficient across all our evaluation axes.

We find that as we move the training region farther from the transition, the empirical transfer coefficient and GEVP bound both grow as a power law in extrapolation distance. The consistency of this behavior across five datasets suggests a pattern. For polynomials, the power-law structure is implicit in the Volume bound in \citet{kalavasis2024transfer} whose relative-volume factors grow with the gap. RFFs have no notion of degree, so it is not obvious that a power law should appear at all. We prove, using the Tur\'an--Nazarov inequality for exponential sums, that RFFs satisfy a corresponding power-law upper bound in distance, with a distance-independent spectral bound set by the feature frequencies and the training-region width. Together, this provides evidence that extrapolation via polynomial and kernel regression across phase boundaries may be feasible and principled.

Our main contributions are as follows.
\begin{enumerate}[leftmargin=*, topsep=-2pt, noitemsep]
    \item We formalize extrapolation across phase transitions as a class of learning problems where the target behavior in the extrapolation regime is dramatically different from that observed during training (\Cref{sec:setup}). Because the change is sharp, structured, and physically interpretable, this setting provides a benchmark for studying principled extrapolation beyond the training regime.
    \item We show that polynomial and kernel regression extrapolate more reliably than the neural architectures we test, which include physics-informed~\citep{raissi2019physics} and spectral-bias variants~\citep{tancik2020fourier, sitzmann2020siren} (\Cref{sec:experiments}).
    \item We upper bound the extrapolation error by using the GEVP to control the worst-case transfer coefficient, a quantity we show to be the tightest bound on the empirical transfer coefficient. The GEVP bound requires no target phase labels and stays approximately constant relative to the observed transfer coefficient. We further observe that, for polynomial and random Fourier features, the extrapolation error grows as a power law in the extrapolation distance, and prove corresponding power-law upper bounds on the transfer coefficient together with a spectral bound that prevents indefinite growth for RFFs (\Cref{sec:methods}, \Cref{sec:experiments}).
\end{enumerate}

\section{Problem Setup}
\label{sec:setup}
We regress a noisy target $y = f^*(x) + \eta$ on inputs $x = (T, s) \in \mathcal T \times \mathcal S \subset \mathbb R^2$, where $f^*: \mathcal T \times \mathcal S \to \mathbb R$. Here $y$ corresponds to a physical observable such as the free energy, $T$ is a tuning parameter such as temperature, and $s$ is a system-specific coordinate.  The training region $P$ lies  on one side of the critical transition $T_c$. The extrapolation region $Q$ extends across $T_c$ and contains the phase of interest. Both $P$ and $Q$ are uniform measures over their regions, which we take to be disjoint rectangles. Let $D$ be the distance from the near edge of $P$ to $T_c$. Increasing $D$ moves the training region farther back from the transition while the extrapolation region stays fixed.

We draw $n$ inputs $x_i \sim P$ and observe $y_i = f^*(x_i) + \eta_i$. We fit each learner $\hat f$ on 80\% of these samples by empirical risk minimization. We report the validation error $\mathrm{MSE}_P(\hat f) = \mathbb{E}_{x \sim P}[(f^*(x) - \hat f(x))^2]$ on the remaining 20\% of inputs and sample $2000$ points from $Q$ to evaluate the extrapolation error $\mathrm{MSE}_Q(\hat f) = \mathbb{E}_{x \sim Q}[(f^*(x) - \hat f(x))^2]$. Hyperparameters are selected based on inputs from $P$ alone, forcing our learner $\hat f$ to strictly learn training region physics.

Let $\mathcal F$ be any function class containing the residual $r = f^* - \hat f$. The worst-case transfer coefficient can be defined as $\Gamma(\mathcal{F}; P, Q) := \sup_{g \in \mathcal{F},\, g \neq 0} \mathbb{E}_{x \sim Q}[g(x)^2] / \mathbb{E}_{x \sim P}[g(x)^2]$. It depends only on the geometry of $P$, $Q$, and the chosen class, which yields $\mathrm{MSE}_Q(\hat f) \le \Gamma(\mathcal{F}; P, Q) \cdot \mathrm{MSE}_P(\hat f)$. This quantity converts a validation MSE in the training region into an extrapolation MSE. Thus, every bound below is simply a particular choice of $\mathcal{F}$ obtained from a computation of $\Gamma$.

\label{sec:setup-physics}
Successful extrapolation should recover both the physical observable structure and the derived nonanalytic quantity, extracted from the predicted surface after fitting. Below, we consider five examples that range from closed-form equations to numerical solutions to experimental measurement. Further dataset generation and preprocessing details are deferred to \Cref{app:datasets}. 

\smallskip

\noindent \textbf{Heisenberg (analytic, polynomial).}
We predict the mean-field Landau--Ginzburg free-energy surface $f(T,m)$ of a ferromagnetic Heisenberg model~\citep{landau1980statistical, vargas2018extrapolating} as a function of temperature $T$ and magnetization $m$. For the paramagnetic $T > T_c$ phase, the free-energy minimum occurs at $m = 0$. For the ferromagnetic $T < T_c$ phase, the surface develops two symmetric minima at nonzero magnetization. The order parameter is the absolute magnetization that minimizes the free energy, $|m^*(T)| = |\arg\min_m f(T,m)|$.

\smallskip

\noindent \textbf{Flory--Huggins (analytic, non-polynomial).} We predict the mixing free energy $f(T,m)$ of a symmetric binary mixture~\citep{flory1942thermodynamics, huggins1941solutions} as a function of temperature $T$ and composition $m = x - 1/2$, where $x$ is the volume fraction of one component. Above $T_c$, entropy favors mixing to yield a single free-energy minimum at equal composition. Below $T_c$, the mixture separates into two phases to produce a symmetric double well.  The order parameter is the magnitude of the minimizing composition relative to equal mixture, $|m^*(T)|$.

\smallskip

\noindent \textbf{SSH Polaron (numerically simulated).} We reproduce \citet{vargas2018extrapolating}, using the dispersion $E(K; \lambda_{\text{SSH}})$ of a single polaron in the Su--Schrieffer--Heeger model~\citep{marchand2010sharp}, with dimensionless electron--phonon coupling $\lambda_{\text{SSH}}$ as the tuning parameter and momentum $K$ as the second coordinate. At weak coupling the dispersion has its minimum at $K=0$ and the ground state carries no momentum. Past a critical coupling $\lambda_c$, the minimum moves continuously to finite $K^*(\lambda) = \arg\min_K E$.\footnote{Technically, this describes a sharp nonanalyticity of the ground state rather than an actual phase transition. For our purposes it behaves mathematically the same as a phase transition with $K^*$ zero on one side and nonzero on the other.}

\smallskip

\noindent \textbf{Rayleigh--B\'enard (numerically simulated).} We predict the mean temperature profile $\bar T(\mathrm{Ra},z)$ of a fluid layer heated from below~\citep{ahlers2009heat}, as a function of the Rayleigh number $\mathrm{Ra}$, which measures the strength of the heating, and height $z$. At low $\mathrm{Ra}$, the fluid is motionless and heat flows by conduction, giving a linear temperature profile $\bar T=1-z$ independent of $\mathrm{Ra}$. Above a critical $\mathrm{Ra}$, convection mixes the interior and flattens the temperature profile by an amount that grows with $\mathrm{Ra}$. The order parameter is the convective amplitude $\Phi=\sqrt{\langle(\bar T-(1-z))^2\rangle_z}$, where $\langle\cdot\rangle_z$ denotes the average over height. It measures departure from conduction, remaining zero below the transition and growing above it.

\smallskip

\noindent \textbf{ARPES (experimental).} We predict the angle-resolved photoemission spectroscopy (ARPES) intensity $I(T,E)$, which probes the electronic spectral function of the overdoped Pb-doped cuprate superconductor Bi-2212, as a function of temperature $T$ and binding energy $E$~\citep{sobota2021angle, he2021rapid}. The measured intensity responds to the superconducting gap, but it does not define the transition. In this sample the antinodal gap closes at $T_{\mathrm{gap}}\approx90$ K, well above the resistive $T_c\approx66$ K. We define the training and extrapolation regions relative to $T_{\mathrm{gap}}$, marked in ARPES panels, and track gap closing through the spectral peak energy $E^*(T)$.

The bound $\mathrm{MSE}_Q(\hat f) \le \Gamma(\mathcal{F}; P, Q) \cdot \mathrm{MSE}_P(\hat f)$ is useful only if $\mathcal{F}$ is expressive enough to contain $r = f^* - \hat f$. The function classes we consider are polynomials, kernels, and neural networks. All are universal approximators in the limit of unbounded degree, RKHS norm, or width, so in principle any of them captures the residual. However, that same universality is why $\Gamma$ is meaningless without restrictions. If $\mathcal{F}$ contains every function, a member can fit the training data exactly while taking arbitrarily large values on $Q$, making $\Gamma$ infinite. Put simply, without restrictions on $\mathcal{F}$, extrapolation across a disjoint domain is ill-posed. We thus restrict $\mathcal F$ to finite-dimensional classes.

The remaining question is which finite class to use. For the free-energy targets, Landau theory~\citep{landau1980statistical} justifies viewing the free energy near a continuous transition as a low-order polynomial expansion about the critical point. This directly motivates a low-degree polynomial class for the analytic Heisenberg model which takes the shape of a quartic polynomial in $m$. For Flory--Huggins, SSH Polaron, Rayleigh--B\'enard, and ARPES, the polynomial class is based on the assumption that each observable, being a smooth function of its inputs, can be approximated by a low-order polynomial, as the physics is likely governed by low-dimensional effective equations. Their residuals therefore lie only approximately in the class. While this assumption is unjustified \emph{a priori}, we find strong extrapolation results and tight GEVP upper bounds on all five targets, suggesting that a restricted feature class captures the dominant structure of each observable. We also study kernel methods following \citet{vargas2018extrapolating} who showed that GPs with compositional kernels extrapolate across phase transitions. A GP's function space, however, is infinite-dimensional, so our finite-feature bound cannot directly apply to it. We therefore study random Fourier features, a finite-dimensional approximation to translation-invariant kernels~\citep{bach2017equivalence, rudi2017generalization}, to which our bound applies. Polynomials and random Fourier features are the two finite classes that we use for bounded extrapolation.

\section{Extrapolation with Bounded Error}
\label{sec:methods}

We compare four learners fit on samples from the training region $P$ and evaluated on the extrapolation region $Q$. Per-method training details and hyperparameter grids are deferred to \Cref{app:training}.

\subsection{Fixed-Feature Regression}

\paragraph{Polynomials.}
\label{sec:methods-poly}

We fit $\hat f$ on the training split using ridge regression with a multivariate polynomial feature expansion of maximum total degree $d$.  Below, we present three bounds for the polynomial residual class $\mathbb P_d$, in increasing order of the extent to which they make use of the specific geometry of $P$ and $Q$. Proofs are found in \Cref{app:derivations}.

\begin{proposition}[Volume bound, \citet{kalavasis2024transfer}]
\label{prop:volume-bound}
Let $P,Q \subset R_{\mathrm{full}} \subset \mathbb{R}^n$ be disjoint rectangles with uniform measure, and define $\rho_P = |R_{\mathrm{full}}|/|P|$ and $\rho_Q = |R_{\mathrm{full}}|/|Q|$. If $r = f^* - \hat f \in \mathbb P_d$, then 
    $\mathrm{MSE}_Q(r)
    \le
    (2d)^{2d}\,\rho_Q\,\rho_P^{2d}\,\mathrm{MSE}_P(r)$.
\end{proposition}
\begin{proposition}[Remez bound, \citet{brudnyi1973}]
\label{prop:remez-bound}
Let $R_{\mathrm{full}} \subset \mathbb{R}^n$ be convex, $P,Q \subseteq R_{\mathrm{full}}$, and $\rho = |R_{\mathrm{full}}| / |P|$, and let $r \in \mathbb{P}_d$ satisfy $\varepsilon = \sup_{x \in P}|r(x)|$. Then
    $\mathrm{MSE}_Q(r)
    \le
    \left(4n\rho\right)^{2d}\varepsilon^2.$
\end{proposition}

Here $|\cdot|$ denotes volume, $R_{\mathrm{full}}$ is the smallest rectangle containing $P$ and $Q$ in our experiments, and $n$ is the input dimension ($n=2$ here), not the number of training samples. Both bounds are geometry-based and apply under disjoint support without any samples or labels in $Q$. They depend on $P$ and $Q$ only through the ratios of their volumes to that of $R_{\mathrm{full}}$, so the same bounds apply to every pair of regions with those ratios, regardless of shape or location. Both bounds are exponentially large in degree $d$ and sit many orders of magnitude above the empirical transfer coefficient in \Cref{fig:bounds-vs-n}. The Remez bound also replaces the training MSE with the maximum squared residual on $P$, making it loose when this maximum greatly exceeds the average training error. 

\begin{proposition}[Finite-feature GEVP bound]
\label{prop:finite-feature-gevp-bound}
Let $\phi:\mathcal X\to\mathbb R^N$ be a fixed feature map and let $\mathcal F_\phi=\{w^\top\phi:w\in\mathbb R^N\}$. For probability measures $P$ and $Q$ on $\mathcal X$, define $A=\mathbb E_{x\sim Q}[\phi(x)\phi(x)^\top]$ and $B=\mathbb E_{x\sim P}[\phi(x)\phi(x)^\top].$ If $B$ is positive definite on the feature span, then
\[
    \Gamma(\mathcal F_\phi;P,Q)
    =
    \sup_{r \in \mathcal F_\phi,\, r \neq 0}
    \frac{\mathbb{E}_{x \sim Q}[r(x)^2]}
         {\mathbb{E}_{x \sim P}[r(x)^2]}
    =
    \lambda_{\max}(A,B).
\]
If $r=f^*-\hat f$ lies in $\mathcal F_\phi$, then $\mathrm{MSE}_Q(\hat f)\le\lambda_{\max}(A,B)\,\mathrm{MSE}_P(\hat f)$.
\end{proposition}
Note that the GEVP bound we develop applies to a broader class of extrapolation problems beyond the ones here. Under the assumptions of \Cref{prop:finite-feature-gevp-bound}, it accommodates any dimension, training and extrapolation regions of arbitrary shape and volume, and uniform or nonuniform sampling.

For degree-$d$ polynomial regression, taking $\phi=\phi_d=(x^\alpha)_{|\alpha|\le d}$ gives $\mathcal F_\phi=\mathbb P_d$ and yields the polynomial GEVP bound through \Cref{prop:finite-feature-gevp-bound}. The expectations defining $A$ and $B$ use the known distributions $P$ and $Q$, so computing them requires no labels from either region. We estimate these matrices from $10{,}000$ unlabeled Monte Carlo draws from each rectangle, using the same budget across extrapolation distances. This choice is motivated by a matrix-Chernoff bound~\citep{tropp2012user} and supported by convergence checks on Heisenberg and ARPES, where both polynomial and RFF estimates remain accurate as the separation increases. \Cref{app:derivations} provides the derivation and results. The GEVP bound identifies the feature-space direction with the greatest error amplification, maximizing the ratio of mean squared values on $Q$ and $P$. Unlike the Volume and Remez bounds, it is computed from the features of $P$ and $Q$ rather than from just their volume ratios, making it far tighter (\Cref{fig:bounds-vs-n}).

\vspace{-3mm}

\paragraph{Gaussian Processes.}
\label{sec:methods-gp}

Gaussian processes are the most expressive kernel baseline for this problem. \citet{vargas2018extrapolating} found that GPs with iteratively built compositional kernels~\citep{duvenaud2013structure} extrapolate well across phase transitions, including the Heisenberg free energy and SSH polaron model that we reuse in our benchmarks. We replicate their GP training protocol, with details in \Cref{app:training}. A GP reports its own uncertainty without labels. Empirically, the posterior variance tracks the shape of the extrapolation error but underestimates its magnitude, so it cannot serve as an upper bound in our experiments (\Cref{app:gp-variance}). On the other hand, the GEVP bound is computable only for finite feature representations, whereas the RKHS of a GP kernel is infinite-dimensional. Fortunately, random Fourier features (RFFs) serve as a finite-dimensional approximation to translation-invariant kernels, so we use them as the finite feature class to which we apply the GEVP bound.

\vspace{-3mm}

\paragraph{Random Fourier Features.}
\label{sec:methods-rff}

Random Fourier features approximate a kernel by replacing it with a finite set of random sinusoidal features~\citep{rahimi2007rff}. In our implementation, we draw $m$ features from a Gaussian spectral measure, $z_j(x) = \sqrt{2/m}\cos(\omega_j^\top x + b_j),\,j = 1,\dots,m$,
with $b_j$ uniform on $[0,2\pi)$, and fit $\hat f$ by ridge regression in the resulting $m$-dimensional feature space.  Once trained, the frequencies and phases are held fixed and the RFF model becomes linear in the feature map $\phi=z$, so \Cref{prop:finite-feature-gevp-bound} applies directly and gives the RFF GEVP bound.

Random-feature approximation motivates using RFFs as a finite-dimensional proxy for kernel methods. Under regularity and sampling assumptions, and when the features are drawn from the spectral measure of the same kernel, the finite random-feature space approximates the kernel RKHS in $L^2$ of the input distribution with comparable statistical guarantees~\citep{bach2017equivalence, rudi2017generalization}. Our features are drawn from a Gaussian spectral measure while the GP uses an iteratively built compositional kernel, so, strictly speaking,  these assumptions do not hold. We therefore treat the RFF GEVP bound as exact for the finite RFF model and as an empirical proxy for the GP. Our experiments show that the RFF empirical transfer coefficient converges toward the GP's empirical transfer coefficient as $m$ grows and tracks it across training-set sizes, validating the use of the RFF GEVP bound as a proxy bound for GPs (\Cref{app:bach}).

\subsection{Learned-Feature Regression}

\paragraph{MLPs.}
\label{sec:methods-mlp}
Multi-layer perceptrons (MLPs) are universal approximators~\citep{cybenko1989} and fit \(f^\ast\) on \(P\) to numerical precision. However, we empirically find MLPs, including several variants such as the Fourier-feature MLP (FF-MLP), SIREN, PINN, and gray-box model, to be incapable of extrapolating successfully across the phase transitions in our datasets. This can be seen easily with the example of ReLU networks, for which predictions approach linear functions along directions away from the training distribution~\citep{xu2021neural}. If \(f^\ast\) changes non-affinely along the extrapolation direction from \(P\) to \(Q\), an affine continuation that persists from the boundary into \(Q\) cannot capture the functional form in \(Q\). Our experiments use GELU and tanh, which impose analogous limitations. GELU networks become asymptotically affine, whereas tanh networks saturate toward a constant. Neither asymptotic form provides a mechanism for inferring an unobserved change in phase structure from the fit on \(P\) alone, because such a change produces a non-affine change in functional form—and therefore a change in slope, curvature, or a higher derivative—between \(P\) and \(Q\). For the continuous transitions considered here, this non-affine change takes the form of a singularity or sharp change in a higher-order derivative. We observe the same limitation empirically in the other architectures that we study. The FF-MLP and SIREN represent the input through sinusoids, so that along any direction into \(Q\) the fitted function is a bounded, quasi-periodic continuation set by frequencies fixed or learned on \(P\). The PINN and gray-box model add physical constraints, but their neural components continue into \(Q\) with the same asymptotic forms discussed above. In other words, although these NN-based models are sufficiently expressive to fit \(f^\ast\) on \(P\), the functions learned on \(P\) do not retain enough flexibility near the boundary and into \(Q\) to recover the more complex phase structure. On the other hand, the fixed-feature regression models discussed above remain non-affine near the boundary between \(P\) and \(Q\) and are thus capable of extending the curvature and other structural information learned within \(P\) into the extrapolation region without being trained on observations from \(Q\).

\section{Empirical Results}
\label{sec:experiments}

\begin{phenomenon}{Fixed-Feature Regression Empirically Extrapolates}
\label{sec:qualitative}
Polynomial, RFF, and GP regression recover the physical structure across the transition on all five datasets. The neural-network methods do not recover the unobserved phase structure.
\end{phenomenon}

\begin{figure}[!ht]
    \centering
    \includegraphics[width=\linewidth]{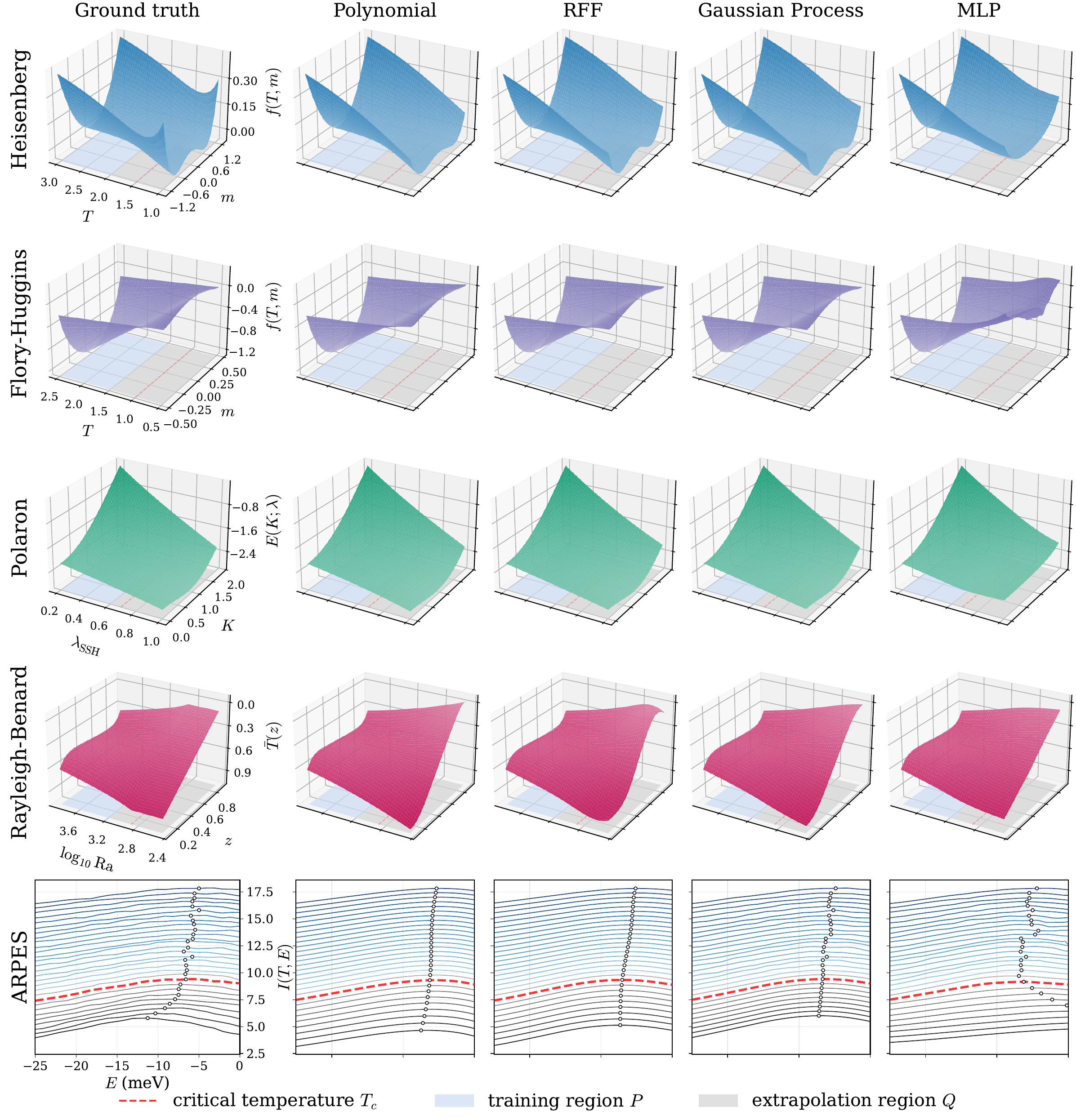}
    \caption{Ground truth and predictions from the four methods for all five targets. For ARPES, the marker tracks the recovered peak energy at each temperature. Polynomial, RFF, and GP regression largely reproduce how each observable changes past the transition, such as the double well that opens in the free-energy targets below $T_c$, while the MLP, on most targets, extends the training-phase trend into $Q$ and fails to capture the qualitative change in behavior.
    }
    \label{fig:landscape}
\end{figure}

\begin{table}[!ht]
    \centering
    \caption{Extrapolation metrics for all five targets. The normalized mean squared error (NMSE) is defined as the mean squared error on $Q$ divided by the variance of the ground truth on $Q$. The relative error is defined as the mean absolute error on $Q$ divided by the standard deviation of the ground truth on $Q$, as a percentage. $|\Delta T_c|/T_c$ is the relative error in the recovered critical point, omitted for ARPES, where the spectral peak responds to the transition but does not define $T_c$.
    }
    \label{tab:metrics}
    \vspace{0.4em}
    \setlength{\tabcolsep}{4pt}
    \begin{tabular}{lccccc}
        \toprule
        & Heisenberg & ARPES & Polaron & Flory--Huggins & Rayleigh--B\'enard \\
        \midrule
        \multicolumn{6}{l}{\textit{Extrapolation NMSE}} \\
        Polynomial & $4.8{\times}10^{-2}$ & $3.8{\times}10^{-2}$ & $3.3{\times}10^{-3}$ & $1.4{\times}10^{-3}$ & $3.2{\times}10^{-2}$ \\
        RFF        & $2.8{\times}10^{-2}$ & $3.4{\times}10^{-2}$ & $3.7{\times}10^{-3}$ & $7.5{\times}10^{-4}$ & $9.2{\times}10^{-2}$ \\
        GP         & $3.3{\times}10^{-2}$ & $8.3{\times}10^{-3}$ & $1.6{\times}10^{-3}$ & $9.0{\times}10^{-8}$ & $5.2{\times}10^{-2}$ \\
        MLP        & $2.0{\times}10^{-1}$ & $1.1{\times}10^{-1}$ & $3.0{\times}10^{-1}$ & $3.2{\times}10^{-1}$ & $1.8{\times}10^{-2}$ \\
        \addlinespace[2pt]
        \multicolumn{6}{l}{\textit{Extrapolation relative error (\%)}} \\
        Polynomial & $7.8$ & $11.9$ & $3.5$ & $2.4$ & $11.2$ \\
        RFF        & $6.5$ & $12.0$ & $3.1$ & $1.7$ & $16.1$ \\
        GP         & $5.5$ & $5.4$ & $2.3$ & $0.0$ & $13.5$ \\
        MLP        & $28.8$ & $23.1$ & $39.5$ & $36.5$ & $9.8$ \\
        \addlinespace[2pt]
        \multicolumn{6}{l}{\textit{Critical-point error $|\Delta T_c| / T_c$ (\%)}} \\
        Polynomial & $4.7$ & --- & $0.7$ & $3.7$ & $5.0$ \\
        RFF        & $1.6$ & --- & $0.4$ & $2.6$ & $2.9$ \\
        GP         & $1.3$ & --- & $0.9$ & $0.4$ & $2.1$ \\
        MLP        & $15.7$ & --- & $14.9$ & $34.2$ & $10.8$ \\
        \bottomrule
    \end{tabular}
\end{table}

\begin{figure}[!ht]
    \centering
    \includegraphics[width=0.8\linewidth]{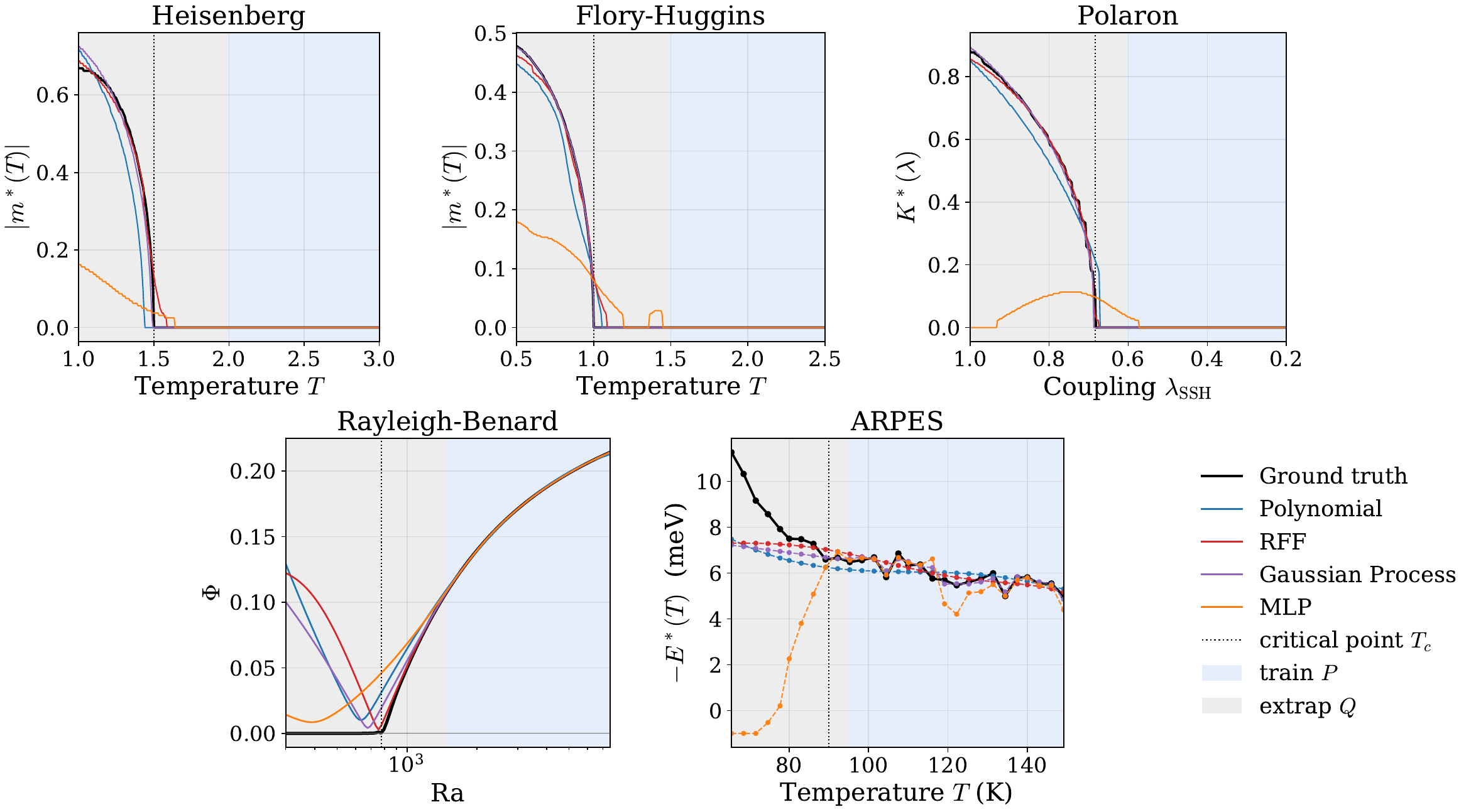}
    \caption{Derived nonanalytic quantity for each target, extracted from the predicted observable surfaces in \Cref{fig:landscape}: $|m^*(T)|$ for Heisenberg and Flory--Huggins, the ground-state momentum $K^*(\lambda)$ for the polaron, the convective amplitude $\Phi$ for Rayleigh--B\'enard, and the peak energy $-E^*(T)$ for ARPES. All quantities are in the natural units of each model: $T$ in units of the model's energy scale with $k_B = 1$ ($T_c = 1.5$ for Heisenberg, $T_c = 1$ for Flory--Huggins), $K$ in units of the inverse lattice constant, and $\lambda_{\mathrm{SSH}}$ and $\mathrm{Ra}$ dimensionless by definition.}
    \label{fig:orderparam}
\end{figure}

Empirical results are shown in \Cref{fig:landscape,fig:orderparam}, where we compare each method's predicted observable and its derived nonanalytic quantity with the ground truth, with extrapolation errors reported in \Cref{tab:metrics}. Across all targets, the MLP fits the training phase to numerical precision, but fails to infer the structure in the unobserved phase. For Heisenberg and Flory--Huggins, it does not learn the double well below $T_c$. For ARPES, the fitted intensity loses the temperature-dependent migration (as denoted by the marker) of the spectral peak across the transition. These failures are clearest in the derived quantities of \Cref{fig:orderparam}. Without an appropriate low-dimensional inductive bias, the MLP does not recover the qualitative structure that appears across the transition. 
Polynomial, RFF, and GP regression do, achieving a lower extrapolation NMSE than the MLP on four of the five targets (\Cref{tab:metrics}) and locating the critical point to within a few percent on available targets.

On Rayleigh--B\'enard, however, the MLP attains the lowest extrapolation error on the observable of any method we test (\Cref{tab:metrics}). Below the transition the profile is $\bar T = 1 - z$ at every Rayleigh number, so the truth in $Q$ does not depend on $\mathrm{Ra}$, whereas the other targets report nonlinear behavior past the transition. Our selected $\tanh$ MLP saturates beyond the data and continues to a nearly $\mathrm{Ra}$-independent profile, so on this target the architecture's default is close to the truth (\Cref{sec:methods-mlp}). Even here, the derived quantity, the convective amplitude $\Phi$, measures departure from that affine profile, so an MLP that extrapolates affinely predicts no departure at any Rayleigh number. Thus, it locates the transition far less accurately than the polynomial and kernel methods (\Cref{fig:orderparam}). Additional fixed-temperature cuts are provided in \Cref{fig:slices}. We find that the four other neural architecture baselines similarly fail to capture this phase change (see \Cref{sec:methods-mlp} and \Cref{app:nn}), demonstrating that this extrapolation failure is not unique to MLPs.
\vspace{3mm}
\begin{phenomenon}{GEVP Tightly Bounds the Extrapolation Error}
\label{sec:sweep}
The GEVP bound upper bounds the observed transfer coefficient across training-set sizes, while the geometry-based bounds sit many orders of magnitude above it. These results extend the setting of \citet{vargas2018extrapolating}, broadening the methods from kernel to polynomial regression, and the targets from analytic and simulated data to noisy experimental measurement. Most importantly, we bound the error before any ground-truth comparison.
\end{phenomenon}

\begin{figure}[!ht]
    \centering
    \includegraphics[width=\linewidth]{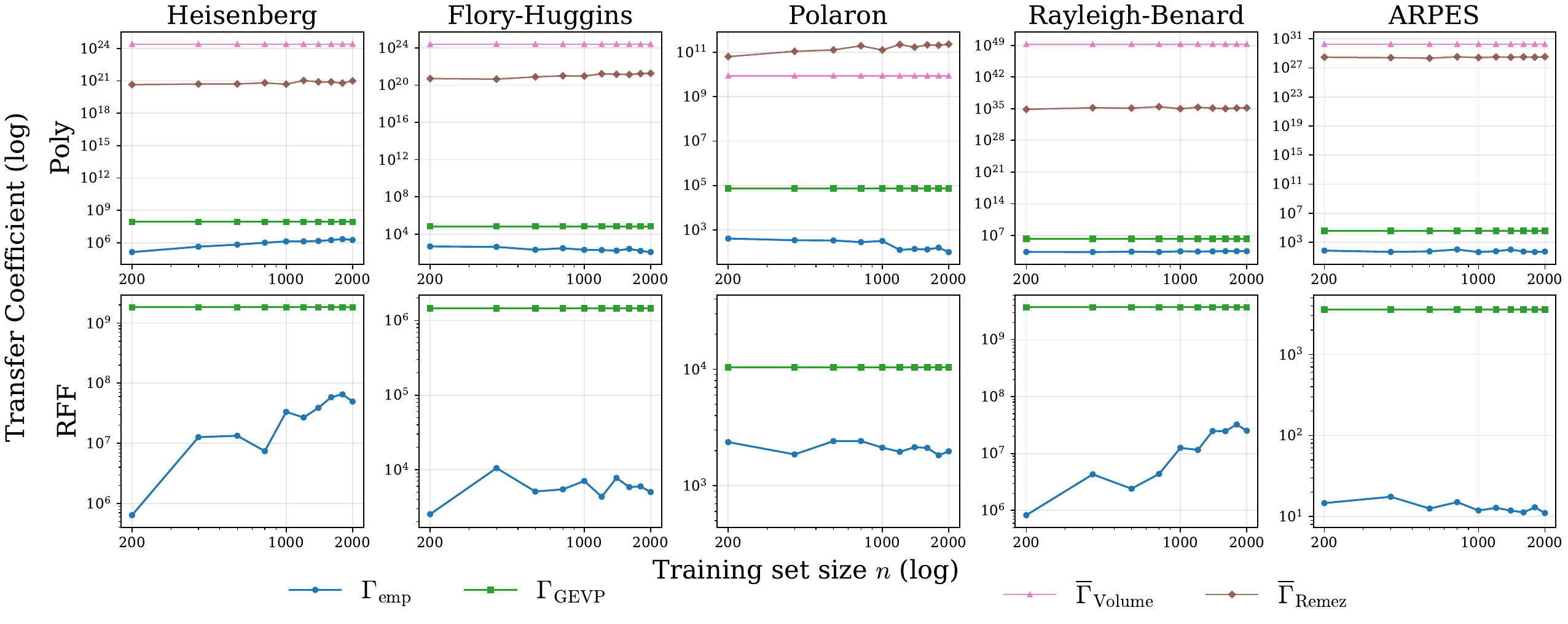}
    \caption{Empirical transfer coefficient $\Gamma_{\mathrm{emp}}$ and the GEVP bound $\Gamma_{\mathrm{GEVP}}$ as the training-set size $n$ varies, for polynomial and RFF regression on all five targets. The top row (Poly) shows both quantities for polynomial regression, together with the Volume and Remez bounds that apply only to polynomials, and the bottom row (RFF) shows them for random Fourier feature regression.
    }
    \label{fig:bounds-vs-n}
\end{figure}

The empirical transfer coefficient depends on the trained model. It changes with the number of labeled training samples $n$, because the validation error on $P$ and the extrapolation error on $Q$ may not fall at the same rate. In contrast, for a fixed feature map the GEVP bound is determined only by the feature covariance geometry of $P$ and $Q$, so it is unchanged across training sizes. As \Cref{fig:bounds-vs-n} shows, for both the polynomial and RFF classes the bound stays above the empirical coefficient, at every training size, at a roughly constant ratio, while the Volume and Remez bounds sit orders of magnitude higher. In other words, the GEVP bound is fixed by geometry, rather than by the amount of data. One can therefore compute an upper bound on their extrapolation error before collecting a single target label. Thus, polynomial and kernel regression may serve as practical methods where the unobserved phase is expensive or impossible to sample, for example in experimental and computational physics research, where access is limited by measurement constraints or simulation costs.

\begin{phenomenon}{Scaling Law of Extrapolation}
Both the empirical transfer coefficient and the GEVP bound exhibit power-law growth over the measured distances, with empirical RFF flattening on some targets. We prove power-law upper bounds for polynomial and RFF classes, together with a spectral bound for RFFs that is independent of distance. These results suggest that extrapolation becomes harder in a structured way rather than catastrophically as the training region moves farther from the transition.
\end{phenomenon}

\begin{figure}[!ht]
    \centering
    \includegraphics[width=\linewidth]{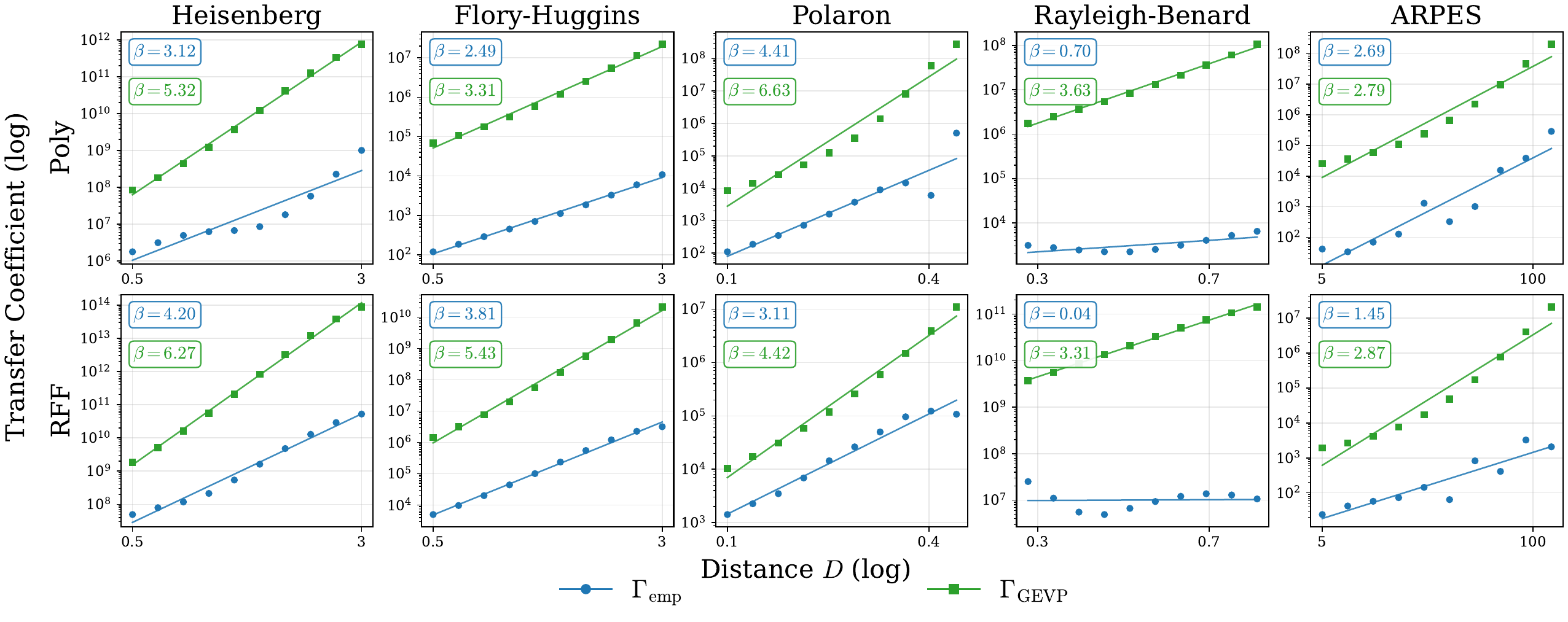}
    \caption{Empirical transfer coefficient $\Gamma_{\mathrm{emp}}$ and GEVP bound $\Gamma_{\mathrm{GEVP}}$ versus extrapolation distance $D$ at fixed training size $n=2000$, for polynomial regression and RFF on all five targets. Several curves exhibit power-law growth with extrapolation distance $D$, where $D$ is the gap between the near edge of the training region and the transition point. For RFFs, the empirical transfer coefficient levels off at larger distances in Polaron and varies little overall in Rayleigh--Bénard. Fitted power-law exponents $\beta$ are annotated in each panel.}
    \label{fig:power-law}
\end{figure}

As the distance $D$ between the training region and the transition grows, extrapolation becomes harder, since the model must learn the implicit phase structure within $P$ well enough to extend it across an increasing gap into $Q$ and capture the phase change. In \Cref{fig:power-law}, the empirical transfer coefficient and GEVP bound exhibit power-law growth over the measured distances, which we dub a scaling law for extrapolation, while the empirical RFF coefficient levels off on some targets. To understand this growth and its limits, we examine how the structure of each hypothesis class controls error amplification with distance.

For polynomials, the scaling law behavior is implicit in the Volume bound (\Cref{prop:volume-bound}). In the distance sweep, the training and extrapolation rectangles keep a fixed volume. Only the enclosing region $R_{\mathrm{full}}$ grows with the gap. The relative volumes $\rho_P$ and $\rho_Q$ grow in proportion to $|R_{\mathrm{full}}|$. The bound $\rho_Q \rho_P^{2d}$ therefore scales as $|R_{\mathrm{full}}|^{2d+1}$, a power law whose exponent is governed by the polynomial degree, and grows with $D$. 

Random Fourier features have no corresponding notion of degree, so a power-law bound requires a different argument. We use the Tur\'an--Nazarov inequality for exponential sums~\citep{nazarov1993local} to establish this bound, then strengthen it with a spectral bound that remains independent of distance. We state the result in one dimension, along the tuning axis.

\begin{theorem}[Power-law and spectral bounds for the RFF transfer coefficient]
\label{thm:rff-power-law}
Fix $m\ge1$ and let $z_j(t)=\sqrt{2/m}\cos(\omega_jt+b_j)$, $j=1,\ldots,m$, be a random Fourier feature map with arbitrary real frequencies and phases, and let $\mathcal G_m=\operatorname{span}_{\mathbb R}\{z_1,\ldots,z_m\}\ne\{0\}$. Let $\lambda_1,\ldots,\lambda_k$ be the distinct elements of $\{\pm\omega_j\}_{j=1}^m$, so $k\le2m$. Let $P$ and $Q$ be disjoint bounded intervals of positive length with uniform measure, and let $I$ be the smallest interval containing both. Define
\[
    \psi(t)=(e^{i\lambda_1t},\ldots,e^{i\lambda_kt})^\top,
    \qquad H_R=\mathbb E_{t\sim R}[\psi(t)\psi(t)^*],\quad R\in\{P,Q\},
\]
where $^*$ denotes the conjugate transpose, and set $\mu_P=\lambda_{\min}(H_P)$ and $\nu_Q=\lambda_{\max}(H_Q)$. Then $\mu_P>0$ and $\nu_Q\le k$. For fixed frequencies, these eigenvalues depend only on the respective interval lengths. There is an absolute constant $C>1$ such that
\[
    \Gamma(\mathcal G_m;P,Q)
    \le \min\left\{
        \left(\frac{C|I|}{|P|}\right)^{2(k-1)},
        \frac{\nu_Q}{\mu_P}
    \right\}.
\]
In particular, $\Gamma(\mathcal G_m;P,Q)\le k/\mu_P$.
\end{theorem}

With the widths of $P$ and $Q$ fixed, the ratio $|I|/|P|$ grows linearly with $D$, so the first term gives a power-law upper bound that holds for any choice of frequencies. This bound accommodates the power-law growth observed over the measured distances, while the second term, the spectral bound, does not depend on $D$ and shows that such growth cannot persist indefinitely for fixed RFF features. The spectral bound depends on the frequency configuration and may be very large. Although it rules out indefinite power-law growth, it does not necessarily predict where the observed growth will slow or flatten. It nevertheless provides a theoretical basis for bounded growth, consistent with the empirical RFF coefficient leveling off beyond $D\approx0.4$ for Polaron and showing little overall growth for Rayleigh--B\'enard. The GEVP bound itself continues to increase over the measured range, however, so these observations do not establish that its spectral limit has been reached. \Cref{app:growth} provides the proof and the reduction from our two-dimensional domains (\Cref{rem:growth-2d}).

\begin{emphasizedremark}[Polynomial degree versus Fourier term count]
An ordinary polynomial can grow through its individual monomials: even the single term $t^d$ exhibits degree-dependent amplification. In a Fourier sum, each atom $e^{i\lambda t}$ has unit modulus, so amplification relative to the training energy arises from cancellations that weaken outside the training region. With $k$ distinct frequencies, a nonzero sum can vanish to order at most $k-1$. The Tur\'an--Nazarov inequality~\citep{nazarov1993local} quantifies this cancellation mechanism, yielding a frequency-uniform squared-error bound with exponent $2(k-1)$. Thus polynomial degree controls the growth available to individual terms, whereas Fourier term count controls cancellation among bounded modes.
\end{emphasizedremark}

The power-law upper bound controls how the extrapolation error grows with the separation between the training region and the transition. For fixed RFF frequencies and training width, the distance-independent spectral bound further limits error amplification at large separations. Together, these bounds and our empirical results provide evidence that extrapolation across phase transitions can be principled and feasible, even when observations near the transition are unavailable.

\section{Related Work}
\label{main:related}

A recent line of work makes extrapolation possible by placing structural assumptions on the data-generating process, such as a causal or noise model~\citep{saengkyongam2024identifying, vonkugelgen2025representation, shen2025engression}, rather than on the hypothesis class as we do. These approaches establish conditions under which the target is identifiable beyond the training support, but do not quantify the extrapolation error. In contrast, our bound quantifies the scale of the extrapolation error, using only the hypothesis class and the training and extrapolation distributions, without explicit labels from the extrapolation region. Transfer-learning theory cannot give such a bound, because it assumes that the training and test distributions overlap~\citep{ma2023optimally, mou2023kernel}. The closest result that allows disjoint support, a bound for polynomials~\citep{kalavasis2024transfer}, is orders of magnitude looser than ours (\Cref{fig:bounds-vs-n}). Within work on phase transitions, \citet{vargas2018extrapolating} showed that Gaussian processes with compositional kernels~\citep{duvenaud2013structure} extrapolate across phase transitions, but without guarantees on the error. Other work on critical transitions forecasts the onset of a transition from data on one side of it~\citep{sevinchan2026incontext, yang2026selfsupervised}, but does not predict the form of the observable on the other side, which is the quantity we extrapolate. Finally, our neural-network results agree with the known failure of these models to extrapolate~\citep{haley1992extrapolation, ziyin2020periodic}, which for ReLU networks takes the form of affine behavior outside the training data~\citep{xu2021neural}. Additional related work is discussed in \Cref{app:related}.

\section{Conclusion}
\label{sec:discussion}

In this work, we study regression methods for extrapolation across continuous transitions, in which a quantity derived from a physical observable diverges or exhibits a kink at the transition point.  We show that polynomial regression and kernel methods, in contrast to the neural network architectures, can extrapolate across a phase boundary and recover the qualitative behavior of the unobserved phase. We demonstrate this on five datasets, spanning analytic free
energies, numerically simulated physical models, and experimental ARPES data for a high-$T_c$ superconductor. To quantify extrapolation reliability, we study the transfer coefficient, which converts validation error on the training region into a bound on error in the extrapolation region. Among the bounds we consider, we show that the GEVP bound we derive bounds the error most tightly because it uses the specific feature-covariance geometry of the training and extrapolation domains, a significant improvement on the Volume and Remez bounds. The bound is computed from the geometry of the two regions alone. It requires no measurement from the unobserved phase, and it is unchanged as the number of labeled samples varies. Most notably, we observe power-law growth in both the empirical transfer coefficient and the GEVP bound over the measured distances, with empirical RFF flattening on some targets. We prove power-law upper bounds for polynomial and random-feature classes, together with a spectral bound that prevents indefinite growth for fixed RFF features, matching our empirical observations. 

This work takes a first step toward a principled framework for extrapolation that can be deployed practically in physics research, for example, when seeking to predict physical behavior in regimes where direct data are limited, inaccessible, or not yet known. As we have shown, a physically motivated finite-dimensional hypothesis class enables accurate empirical extrapolation across continuous transitions, with a worst-case transfer coefficient that i) is bounded from above by quantities that can be estimated from the geometry of the training and extrapolation regions alone, without dependence on the number of labeled samples, and ii) grows at most as a power law in the distance from the training region to the transition. Thus, a practitioner can use the geometry of the problem to estimate the extrapolation error budget beforehand and then deploy the method with confidence that its errors will fall within their accepted tolerance.

\vspace{-3mm}

\paragraph{Limitations.}
\label{sec:limitations}
The main limitation of the GEVP approach is that it requires the residual $r=f^*-\hat f$ to lie in a specified finite-dimensional hypothesis class, and only in the Heisenberg case does that class contain $f^*$ exactly. On the other four targets the class is misspecified. We instead justify this choice by the smoothness of the observable away from the critical point. The bound holds empirically on all five, justifying the GEVP approach for our test cases. The power-law upper bounds do not determine the smaller exponents observed empirically, and the spectral bound for RFFs does not necessarily predict where growth will slow or flatten. Determining the observed exponents and when flattening begins remains an open problem requiring more specialized mathematical methods. The GEVP bound is formalized for the finite RFF model, but its use as a proxy for the GP is empirical, since our Gaussian RFF features do not directly approximate the iteratively constructed compositional GP kernel. Finally, our experiments focus on continuous transitions. While we expect that expert physical constraints may allow similar extrapolation bounds to be useful in other physics settings, extending them to other phenomena remains to be tested in future work.

\section*{Acknowledgments}
This research was supported in part by
the Yale Office of the Provost and the Yale College Dean's Research Fellowship. We acknowledge useful discussions with Wei Bu and Arthur Gretton. We thank Yu He and Jinming Yang for providing the Pb-doped Bi-2212 ARPES dataset used in this work.

\bibliographystyle{unsrtnat}
\bibliography{references}

\newpage
\appendix

\makeatletter
\let\addcontentsline\tocentry
\makeatother

\section*{Appendix}

\vspace{-0.5em}
{\parskip=0pt\renewcommand{\contentsname}{\vspace{-1.5em}}\tableofcontents}
\vspace{0.5em}
\clearpage
\section{Additional Related Work}
\label{app:related}

\paragraph{Machine learning for phase transitions.}
Supervised and unsupervised models have been used to detect, classify, or locate phases from simulation and experimental data~\citep{carrasquilla2017machine, vannieuwenburg2017learning, wang2016discovering, wetzel2017unsupervised}. Closest to our setting, \citet{chen2025detecting} classify the superconducting phase from single ARPES snapshots of the same Bi-2212 measurements we use here. We instead study whether a model trained in one phase can extrapolate the observable itself into an unobserved phase. \citet{vargas2018extrapolating} showed that Gaussian processes with iteratively constructed compositional kernels can do so, building on automatic kernel construction~\citep{grosse2013bayes, duvenaud2013structure, lloyd2014automatic, wilson2013gaussian}, but without a theoretical account of when such extrapolation succeeds or how large its error can be.

\vspace{-3mm}

\paragraph{Neural network extrapolation.}
\citet{haley1992extrapolation} documented that feedforward networks fail to extrapolate smooth functions beyond the training distribution. \citet{xu2021neural} prove that ReLU networks become affine outside the training hull, \citet{ziyin2020periodic} that standard activations cannot extrapolate periodic targets, and \citet{trask2018nalu} that networks do not generalize to unseen numerical ranges. These failures are commonly attributed to a spectral and simplicity bias toward low-frequency, near-affine solutions~\citep{rahaman2019spectral, shah2020pitfalls}. That bias motivates the Fourier-feature and sinusoidal variants we test~\citep{tancik2020fourier, sitzmann2020siren}. Physics-informed neural networks also extrapolate poorly beyond the training domain, even with the governing equations in the loss~\citep{kim2021dpm, krishnapriyan2021characterizing, fesser2023extrapolation}. Models that hard-code the functional form extrapolate better, such as equation learners~\citep{martius2016extrapolation, sahoo2018learning}, symbolic distillation~\citep{cranmer2020discovering}, and the symbolic physical models of \citet{dakhmouche2025why}. This is consistent with our finding that inductive bias rather than in-distribution fit governs extrapolation. Theoretical treatments of distribution shift bound the target error through a divergence or support-coverage term that requires overlap~\citep{bendavid2010theory, mansour2009domain, johansson2019support}. \citet{balestriero2021learning} note that in high dimensions the interpolation/extrapolation distinction stops tracking generalization. Neither assumption holds in our disjoint, low-dimensional setting.

\vspace{-3mm}

\paragraph{Extrapolation beyond the training support.}
A recent line of work extrapolates by restricting the data-generating process rather than the hypothesis class. \citet{saengkyongam2024identifying} and \citet{vonkugelgen2025representation} predict a system's behavior under unobserved experimental conditions, assuming that the effect of each condition combines linearly or additively in a hidden representation, so that an unseen condition is a combination of seen ones. Both extrapolate to new discrete conditions, whereas we extrapolate along a continuous parameter into an unobserved phase. \citet{shen2025engression} fit the full conditional distribution rather than its mean, which constrains the function just past the edge of the training support when the target is monotone. Our target is neither adjacent to the training data nor monotone, as it lies across a gap, in a double well. These works characterize when extrapolation is possible or propose methods for it, whereas our main contribution is a sample-free upper bound on the extrapolation error.

\vspace{-3mm}

\paragraph{Transfer under disjoint support.}
Disjoint support between the training and extrapolation regions falls outside most of transfer-learning theory. Minimax results for RKHS regression under covariate shift are stated in terms of the likelihood ratio between source and target~\citep{ma2023optimally}, and the guarantees obtained without strict overlap still assume the training distribution covers the input space~\citep{mou2023kernel}; neither applies when $P$ and $Q$ are disjoint. The closest result is the non-overlap polynomial transfer bound of \citet{kalavasis2024transfer}. We compare it to a multivariate Remez bound~\citep{brudnyi1973} and to our GEVP bound, which tightens both by using the feature covariances of $P$ and $Q$.

\vspace{-3mm}

\paragraph{Critical transitions and tipping points.}
A complementary line of work predicts when a critical transition occurs from observed dynamics, by in-context learning~\citep{sevinchan2026incontext} or self-supervised reservoir computing~\citep{yang2026selfsupervised}. These methods locate the transition without predicting the observable beyond it or bounding the error. Conformal methods go one step further, maintaining coverage across a change point in a time series~\citep{sun2025conformal}, but they calibrate on labeled points observed after the change, which is exactly what is unavailable here. We address the converse question, and locating the transition then follows as a downstream task. We recover $T_c$ from the extrapolated surface without ever fitting to it.

\section{Proofs and Estimations of the Bounds}
\label{app:derivations}

\paragraph{Volume bound.}
We defer the proof of \Cref{prop:volume-bound} to Theorem 2.3 from \citet{kalavasis2024transfer}.

\vspace{-3mm}

\paragraph{Remez bound.} \Cref{prop:remez-bound} follows from the multivariate Remez inequality~\citep{brudnyi1973}, which controls the maximum of a polynomial on a convex set by its maximum on a subset. With $R_{\mathrm{full}} \subset \mathbb{R}^n$ convex, $P \subset R_{\mathrm{full}}$ of relative measure $\rho = |R_{\mathrm{full}}| / |P|$, and $r \in \mathbb{P}_d$,
\begin{equation*}
    \sup_{x \in R_{\mathrm{full}}} |r(x)|
    \;\le\;
    T_d\!\left(\frac{1 + (1 - 1/\rho)^{1/n}}{1 - (1 - 1/\rho)^{1/n}}\right)
    \sup_{x \in P} |r(x)|,
\end{equation*}
where $T_d$ is the degree-$d$ Chebyshev polynomial, which is extremal for the one-dimensional problem, evaluated at an argument that grows as $P$ occupies less of $R_{\mathrm{full}}$. \citet{brudnyi1973} also bound the Chebyshev factor by $(4n\rho)^{d}$, which gives the form,
\begin{equation*}
    \sup_{x \in R_{\mathrm{full}}} |r(x)|
    \;\le\;
    \left(4n\rho\right)^{d} \sup_{x \in P} |r(x)| .
\end{equation*}
Squaring yields the pointwise bound $r(x)^2 \le (4n\rho)^{2d}\,\varepsilon^2$ on all of $R_{\mathrm{full}}$, with $\varepsilon = \sup_{x \in P}|r(x)|$, and averaging over $Q \subset R_{\mathrm{full}}$ yields \Cref{prop:remez-bound}. 

\vspace{-3mm}

\paragraph{GEVP bound.}
Let $\phi$ be a finite-dimensional feature map, $g_w(x) = w^\top\phi(x)$ an element of its span. Let $A$ and $B$ be the second-moment matrices of $\phi$ over $Q$ and $P$ as in \Cref{prop:finite-feature-gevp-bound}. The second moments of $g_w$ are quadratic forms in the coefficients,
\begin{equation*}
    \mathbb{E}_{x \sim P}[g_w(x)^2] = w^\top Bw,
    \qquad
    \mathbb{E}_{x \sim Q}[g_w(x)^2] = w^\top Aw.
\end{equation*}
The transfer ratio of each feature-space direction is a generalized Rayleigh quotient $w^\top A w / w^\top B w$. When $B$ is positive definite, the Rayleigh--Ritz principle gives \begin{equation*}
    \sup_{w \neq 0} \frac{w^\top A w}{w^\top B w}
    =
    \lambda_{\max}(A,B),
\end{equation*}
attained by the generalized eigenvector $v$ satisfying $Av = \lambda_{\max} B v$. Taking $\phi = \phi_d$ and $\phi = z$ yields the polynomial and RFF GEVP bounds, respectively. In practice $B$ is nearly singular, since high-degree monomials and wide RFF maps are strongly correlated on a narrow training window. We therefore compute $\lambda_{\max}(A, B + \epsilon I)$ via \texttt{scipy.linalg.eigh}, with a small jitter $\epsilon$ that keeps the nearly null directions of $B$ from destabilizing the eigensolver.

\vspace{-3mm}

\paragraph{Estimating $A$ and $B$.}
The unlabeled sampling budget quoted in \Cref{sec:methods} comes from a matrix-Chernoff bound~\citep{tropp2012user}. For either region $R\in\{P,Q\}$, let
\begin{equation*}
    M = \mathbb{E}_{x \sim R}\!\left[\phi(x)\phi(x)^\top\right],
    \qquad
    \widehat{M} = \frac{1}{N_{\mathrm{MC}}}\sum_{i=1}^{N_{\mathrm{MC}}} \phi(x_i)\phi(x_i)^\top
\end{equation*}
be the matrix to be estimated and its empirical average over $N_{\mathrm{MC}}$ i.i.d.\ draws $x_i \sim R$. Suppose the leverage bound
\begin{equation*}
    \phi(x)^\top M^{\dagger}\phi(x) \;\le\; \ell
\end{equation*}
holds almost surely. Whitening with respect to $M$ turns the summands into independent positive-semidefinite matrices with identity expectation and maximum eigenvalue at most $\ell$. The matrix-Chernoff inequality~\citep[Corollary 5.2]{tropp2012user} then gives
\begin{equation}
    \Pr\!\left[(1-\eta)M \preceq \widehat{M} \preceq (1+\eta)M\right]
    \;\ge\; 1 - 2\,\mathrm{rank}(M)\exp\!\left(-\frac{N_{\mathrm{MC}}\eta^{2}}{3\ell}\right).
\label{eq:matrix-chernoff}
\end{equation}
The multiplicative approximation therefore holds with probability at least $1-\delta$ once
\begin{equation*}
    N_{\mathrm{MC}} \;\ge\; \frac{3\ell}{\eta^{2}}\log\frac{2\,\mathrm{rank}(M)}{\delta}.
\end{equation*}
When \Cref{eq:matrix-chernoff} holds for both $A$ and $B$, every Rayleigh quotient is preserved to within a factor $(1+\eta)/(1-\eta)$, and hence so is $\lambda_{\max}$. The requirement involves the leverage and rank of the feature map, but not the location of $Q$ or the number of labeled samples. The convergence experiments below support using the same unlabeled sampling budget across the distance sweep.

\begin{figure}[!h]
    \centering
    \includegraphics[width=0.8\linewidth]{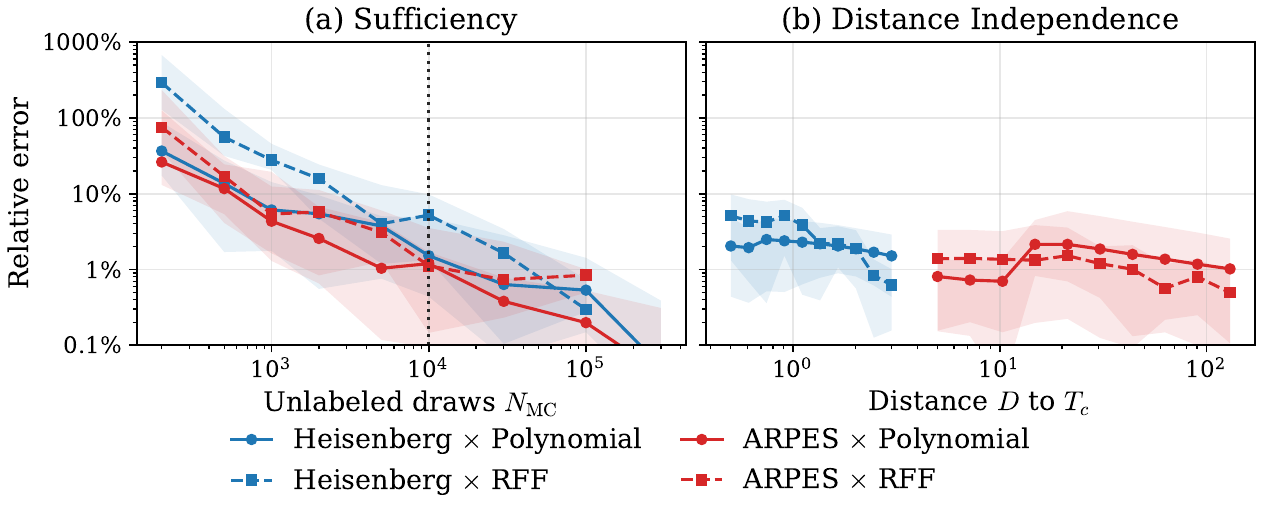}
    \caption{Monte Carlo convergence of the GEVP coefficient. (a) Relative error against a $10^{7}$-draw reference as the number of unlabeled draws $N_{\mathrm{MC}}$ varies. (b) Relative error at $N_{\mathrm{MC}} = 10^{4}$ across the distance sweep. Curves are means over 10 repetitions with 5--95 percentile bands.} \label{fig:gevp-convergence}
\end{figure}

\Cref{fig:gevp-convergence} checks the $N_{\mathrm{MC}} = 10^{4}$ working budget empirically on Heisenberg and ARPES. A reference is computed from roughly $10^{7}$ draws, where \Cref{eq:matrix-chernoff} places the reference error at a few percent. In every dataset $\times$ method, the relative error falls below ten percent well before $10^{4}$ draws. At the working budget, it stays flat as the training region slides, as the bound predicts. 

\section{Power-Law Scaling of the Transfer Coefficient with Distance}
\label{app:growth} We prove \Cref{thm:rff-power-law}. The argument is in one dimension, matching the distance sweep. \Cref{rem:growth-2d} reduces the paper's two-dimensional domains to this case.

\begin{lemma}[{Tur\'an--Nazarov inequality, \citet{nazarov1993local}}]
\label{lem:nazarov}
There is an absolute constant $C_0 > 1$ with the following property. Let
$p(t) = \sum_{j=1}^{k} c_j e^{i \lambda_j t}$ with $c_j \in \mathbb{C}$ and
distinct real frequencies $\lambda_j$, let $I \subset \mathbb{R}$ be an interval,
and let $E \subseteq I$ be measurable with $|E| > 0$. Then
\[
\sup_{I} |p| \;\le\;\Bigl(\frac{C_0\,|I|}{|E|}\Bigr)^{k-1} \sup_{E} |p| .
\]
\end{lemma}

For a subinterval $E$ this is Tur\'an's lemma. \citet{nazarov1993local} extends the lemma to arbitrary measurable sets with an absolute constant. \Cref{lem:nazarov} controls sup norms, while $\Gamma$ is defined through $L^2$ averages. The next lemma bounds the sup norm on $P$ by the $L^2$ average.

\begin{lemma}[Nikolskii-type inequality for exponential sums]
\label{lem:nikolskii}
Let $p$ be as in \Cref{lem:nazarov} with $k \ge 1$ frequencies and let $P$ be an
interval of positive length. Then
$$\sup_{P} |p|^2 \le 2\,(2C_0)^{2(k-1)}\, \mathbb{E}_{t \sim P}[|p(t)|^2],$$ with the same absolute constant $C_0$.
\end{lemma}

\begin{proof}
If $k = 1$, then $|p| \equiv |c_1|$ is constant, so $\sup_P |p|^2 = \mathbb{E}_{t \sim P}[|p(t)|^2]$ and the bound holds trivially. Therefore, we
assume $k \ge 2$. We also assume $p \not\equiv 0$. Then $s := \sup_P |p| > 0$, because an exponential sum is entire and so cannot vanish on an interval unless it vanishes identically.

For $\theta \in (0,1)$ let $E_\theta = \{ t \in P : |p(t)| \le \theta s \}$ be the
set on which $|p|$ falls below the fraction $\theta$ of its peak. If
$|E_\theta| > 0$ then $\sup_{E_\theta} |p| \le \theta s$ by construction.
\Cref{lem:nazarov} applied on $P$ with $E = E_\theta$ yields
\[
s \;\le\; \Bigl(\frac{C_0\,|P|}{|E_\theta|}\Bigr)^{k-1} \theta s
\qquad\Longrightarrow\qquad
|E_\theta| \;\le\; C_0\,|P|\,\theta^{1/(k-1)},
\]
the right-hand inequality holding trivially when $|E_\theta| = 0$. Choosing
$\theta^* = (2C_0)^{-(k-1)}$ makes $(\theta^*)^{1/(k-1)} = 1/(2C_0)$ and hence
$|E_{\theta^\ast}| \le |P|/2$. On the complement, a subset of $P$ of measure at
least $|P|/2$, we have $|p| > \theta^\ast s$, so
\[
\mathbb{E}_{t \sim P}\bigl[|p(t)|^2\bigr]
 \;=\; \frac{1}{|P|} \int_P |p|^2
 \;\ge\; \frac{1}{2}\,(\theta^\ast)^2 s^2
 \;=\; \frac{s^2}{2\,(2C_0)^{2(k-1)}}. \qedhere
\]
\end{proof}

\begin{proof}[Proof of \Cref{thm:rff-power-law}]
Let $g\in\mathcal G_m$ be nonzero. Expanding each cosine into two complex exponentials writes $g(t)=c^*\psi(t)$ for some $c\in\mathbb C^k$, possibly with zero entries. We first prove the power-law bound. If $k=1$, every nonzero function in the exponential span has constant modulus, so its transfer ratio is one and the power-law bound holds. For $k\ge2$, chaining three inequalities gives
\begin{align}
\mathbb{E}_{t \sim Q}[g^2]
 \;&\le\; \sup_Q g^2 \;\le\; \sup_I g^2
 &&
 \notag\\[2pt]
 \;&\le\; \Bigl( \frac{C_0\,|I|}{|P|} \Bigr)^{2(k-1)} \sup_P g^2
 && \text{(\Cref{lem:nazarov})}
 \notag\\[2pt]
 \;&\le\; 2\,(2C_0)^{2(k-1)} \Bigl( \frac{C_0\,|I|}{|P|} \Bigr)^{2(k-1)}
 \mathbb{E}_{t \sim P}[g^2]
 && \text{(\Cref{lem:nikolskii}).}
 \notag
\end{align}
Taking $C=2\sqrt{2}\,C_0^2$, we have $2(2C_0^2)^{2(k-1)}\le C^{2(k-1)}$ for $k\ge2$. Thus,
\[
\frac{\mathbb E_Q[g^2]}{\mathbb E_P[g^2]}
\le \left(\frac{C|I|}{|P|}\right)^{2(k-1)}.
\]
Taking the supremum over $g$ proves the power-law bound.

We next prove the distance-independent bound. Distinct exponentials are linearly independent on every interval of positive length. Indeed, if a sum vanishes on such an interval, its derivatives of orders $0,\ldots,k-1$ vanish at an interior point. The resulting Vandermonde system in the distinct frequencies forces all coefficients to be zero. Consequently, for every nonzero $c\in\mathbb C^k$,
\[
    c^*H_Pc=\mathbb E_{t\sim P}[|c^*\psi(t)|^2]>0,
\]
so $\mu_P>0$. With $g(t)=c^*\psi(t)$, the two second-moment matrices give
\[
    \mathbb E_P[g^2]=c^*H_Pc\ge\mu_P\|c\|_2^2,
    \qquad
    \mathbb E_Q[g^2]=c^*H_Qc\le\nu_Q\|c\|_2^2.
\]
Dividing and taking the supremum over nonzero $g$ yields $\Gamma(\mathcal G_m;P,Q)\le\nu_Q/\mu_P$. Since $H_Q$ is positive semidefinite and $\operatorname{tr}H_Q=k$, we also have $\nu_Q\le k$.

To see how these eigenvalues depend on the intervals, write $\mathcal L=\{\lambda_1,\ldots,\lambda_k\}$ and define
\[
 S_{\mathcal L}(L)=\left[\operatorname{sinc}\!\left(\frac{(\lambda_r-\lambda_s)L}{2}\right)\right]_{r,s=1}^k,
 \qquad
 \operatorname{sinc}(u)=\begin{cases}\sin(u)/u,&u\ne0,\\1,&u=0.\end{cases}
\]
This is the Gram matrix of the exponentials under the uniform probability measure on $[-L/2,L/2]$. If $R$ has midpoint $a$ and length $L$, direct integration gives
\[
 H_R=U_a S_{\mathcal L}(L)U_a^*,
 \qquad U_a=\operatorname{diag}(e^{i\lambda_1a},\ldots,e^{i\lambda_ka}).
\]
The matrix $U_a$ is unitary, so $H_R$ and $S_{\mathcal L}(L)$ have the same eigenvalues. In particular,
\[
 \frac{\nu_Q}{\mu_P}
 =\frac{\lambda_{\max}(S_{\mathcal L}(|Q|))}{\lambda_{\min}(S_{\mathcal L}(|P|))}.
\]
The spectral bound is therefore uniform over interval locations and separations for fixed frequencies and interval lengths. Both bounds hold for every nonzero $g\in\mathcal G_m$, so their minimum bounds the transfer coefficient.
\end{proof}

\begin{remark}[Reduction from the two-dimensional domains]
\label{rem:growth-2d}
Let $P = P_T \times Y$ and $Q = Q_T \times Y$ be the training and
extrapolation rectangles, sharing a transverse window $Y$, with the uniform measures of \Cref{sec:setup}. These are product
measures with the same transverse marginal. Let $I$ be the smallest interval containing $P_T$ and $Q_T$. Fix $y \in Y$. A two-dimensional feature $\sqrt{2/m}\cos(\omega_j^{(1)} t + \omega_j^{(2)} y + b_j)$ is, as a function of $t$ alone, a sinusoid at frequency $\omega_j^{(1)}$ with the transverse part absorbed into its phase. Every $g \in \mathcal{G}_m$ restricts to an exponential sum in $t$ whose frequencies are drawn from $\{\pm\omega_j^{(1)}\}$; let $k \le 2m$ be the number of distinct such frequencies, which does not depend on $y$. Applying \Cref{thm:rff-power-law} at each fixed $y$ and averaging over $y$,
\[
\mathbb{E}_{x \sim Q}[g^2]
 \;=\; \mathbb{E}_{y}\,\mathbb{E}_{t \sim Q_T}[g(t,y)^2]
  \;\le\; \Lambda \, \mathbb{E}_{y}\,\mathbb{E}_{t \sim P_T}[g(t,y)^2]
 \;=\; \Lambda\, \mathbb{E}_{x \sim P}[g^2],
\]
with $\Lambda = \left(C|I|/|P_T|\right)^{2(k-1)}$ independent of $y$, so the
two-dimensional transfer coefficient obeys the same bound.

The spectral bound applies with the same projected frequencies. Define $H_{P_T}$, $H_{Q_T}$, $\mu_{P_T}$, and $\nu_{Q_T}$ as in \Cref{thm:rff-power-law}, using $P_T$ and $Q_T$ in place of $P$ and $Q$. Every slice $g(\cdot,y)$ belongs to this exponential span, so
\[
    \mathbb E_{t\sim Q_T}[g(t,y)^2]
    \le \frac{\nu_{Q_T}}{\mu_{P_T}}\,\mathbb E_{t\sim P_T}[g(t,y)^2].
\]
This inequality also holds when the slice vanishes identically. Averaging over the common transverse marginal bounds the two-dimensional transfer coefficient by $\min\{\Lambda,\nu_{Q_T}/\mu_{P_T}\}$. Since $\nu_{Q_T}\le k$, the bound $k/\mu_{P_T}$ also follows. Translating either rectangle along the extrapolation direction leaves these eigenvalues unchanged.
\end{remark}

\section{RFFs as a Finite-Dimensional Proxy for GPs}
\label{app:bach}

For a continuous translation-invariant positive-definite kernel
$k(x,y) = \kappa(x-y)$ on $\mathbb{R}^n$, Bochner's theorem represents
$\kappa$ as the Fourier transform of a nonnegative finite spectral measure
$\tau$,
\[
    \kappa(x-y)
    =
    \int_{\mathbb{R}^n} e^{i\omega^\top(x-y)} \,d\tau(\omega).
\]
Random Fourier features approximate this integral by Monte Carlo:
drawing $\omega_j \sim \tau$ and $b_j \sim \mathrm{Uniform}[0,2\pi)$ gives the features of \Cref{sec:methods-rff}, with
$\mathbb{E}[z(x)^\top z(y)] = k(x,y)$~\citep{rahimi2007rff}.

The transfer coefficient is well defined only after truncation to $m$ features. For a sufficiently rich RKHS and disjoint regions, the RKHS contains functions arbitrarily small on $P$ yet large on $Q$, so the ratio $\sup_f \mathbb{E}_{x \sim Q}[f(x)^2] / \mathbb{E}_{x \sim P}[f(x)^2]$ is unbounded. In operator terms, the covariance operator of $P$ is compact with no bounded inverse. The generalized eigenvalue problem has no infinite-dimensional analogue. Restricting to the span of $m$ features yields a bound through \Cref{prop:finite-feature-gevp-bound}.

\begin{figure}[!t]
    \centering
    \includegraphics[width=0.8\linewidth]{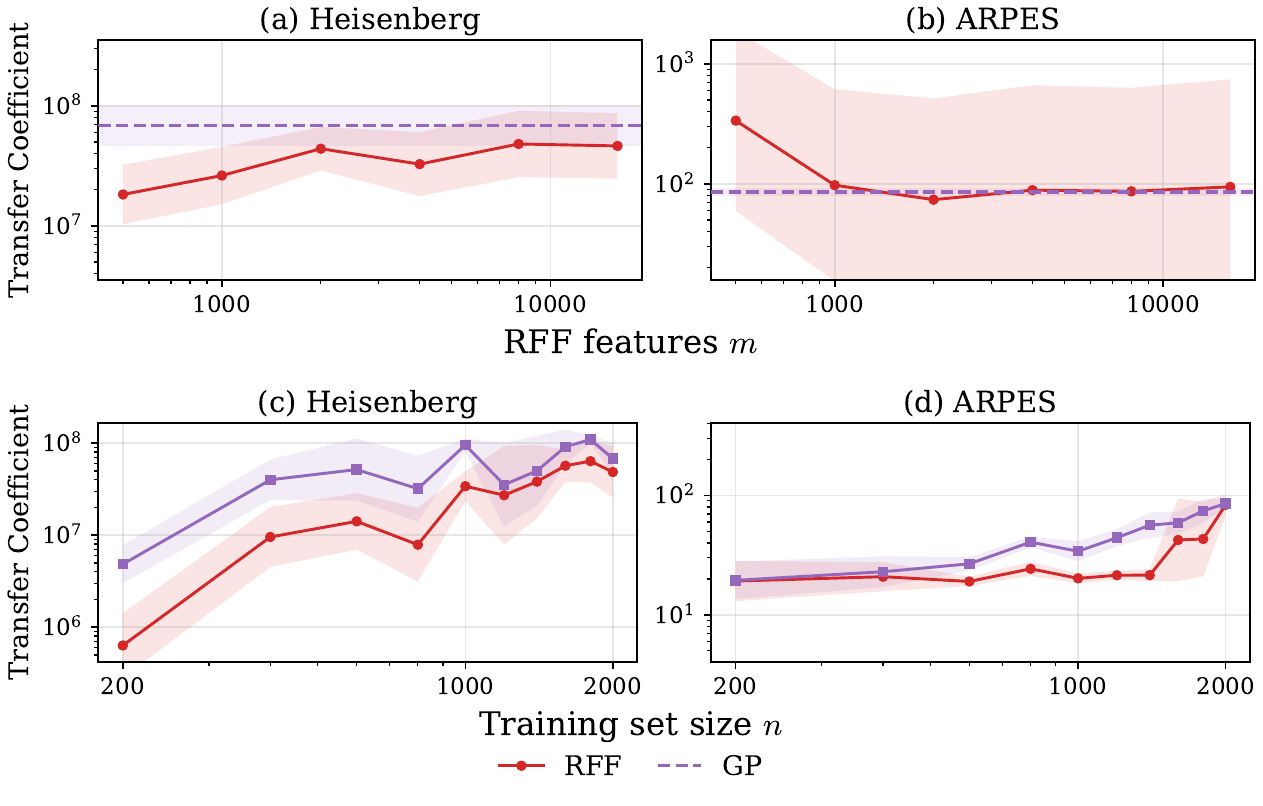}
    \caption{Convergence of the RFF empirical transfer coefficient toward the GP empirical transfer coefficient. Panels (a) and (b) vary the number of random features $m$ at fixed $n = 2000$. The GP value is shown as a dashed line. Panels (c) and (d) vary the training-set size $n$ at fixed $m$. Curves show averages across 5 seeds, and shaded regions indicate $\pm 1$ standard deviation.}
    \label{fig:bach-tracking}
\end{figure}

\Cref{fig:bach-tracking} supports the proxy claim in \Cref{sec:methods-rff} for Heisenberg and ARPES datasets. At a fixed training size, the RFF empirical transfer coefficient converges toward the GP value as $m$ grows. At a fixed number of features $m$, the RFF value tracks the GP value across training-set size.

\section{Alternatives to the GEVP Bound}
\label{app:gp-variance}

Is there anything simpler than the GEVP bound? A Gaussian process already quantifies its uncertainty. Its posterior variance $\sigma^2(x)$ estimates the squared error of the posterior mean $\mu(x)$ pointwise, so its average over $Q$ predicts $\mathrm{MSE}_Q$ without referencing the target phase. Or, one could take a handful of measurements inside the target phase and estimate $\mathrm{MSE}_Q$ directly. We test both on Heisenberg and ARPES, using the selected models at $n = 2000$ and averaging over five seeds.

\begin{figure}[!ht]
    \centering
    \includegraphics[width=0.8\linewidth]{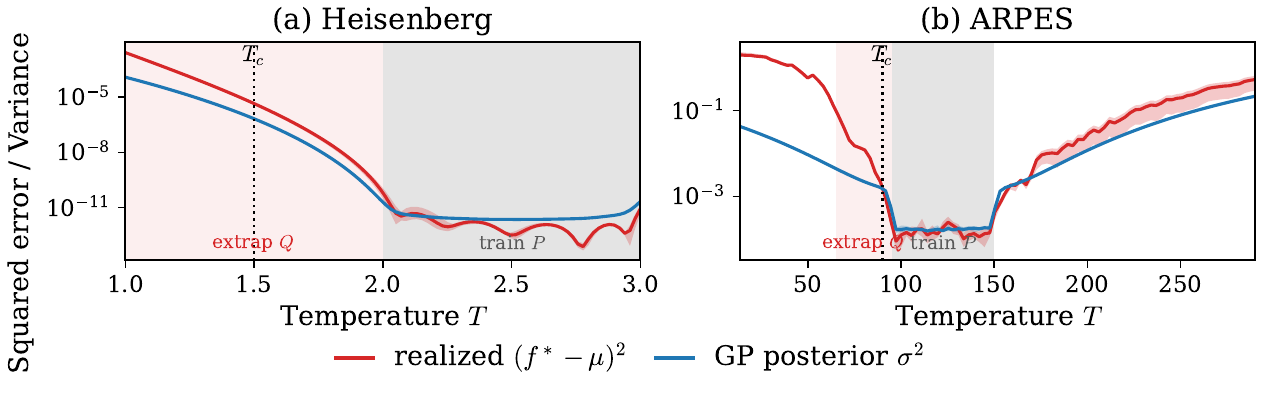}
    \caption{Realized squared error $(f^* - \mu)^2$ and GP posterior variance $\sigma^2$
    along the temperature axis, averaged over the second coordinate and over 5 seeds.}
    \label{fig:gp-variance}
\end{figure}

\Cref{fig:gp-variance} shows the two curves tracing one another inside $P$ and pulling apart below the transition. Inside the training region the posterior variance is a faithful estimate of error. Almost every point falls within two standard deviations of the prediction. Below the transition the variance sinks beneath the error, by about an order of magnitude on Heisenberg and a factor of five on ARPES. Up to an overall scale factor, the variance still tracks the error. The posterior variance therefore locates where the predictions fail but does not estimate $\mathrm{MSE}_Q$.

\begin{figure}[!ht]
    \centering
    \includegraphics[width=0.8\linewidth]{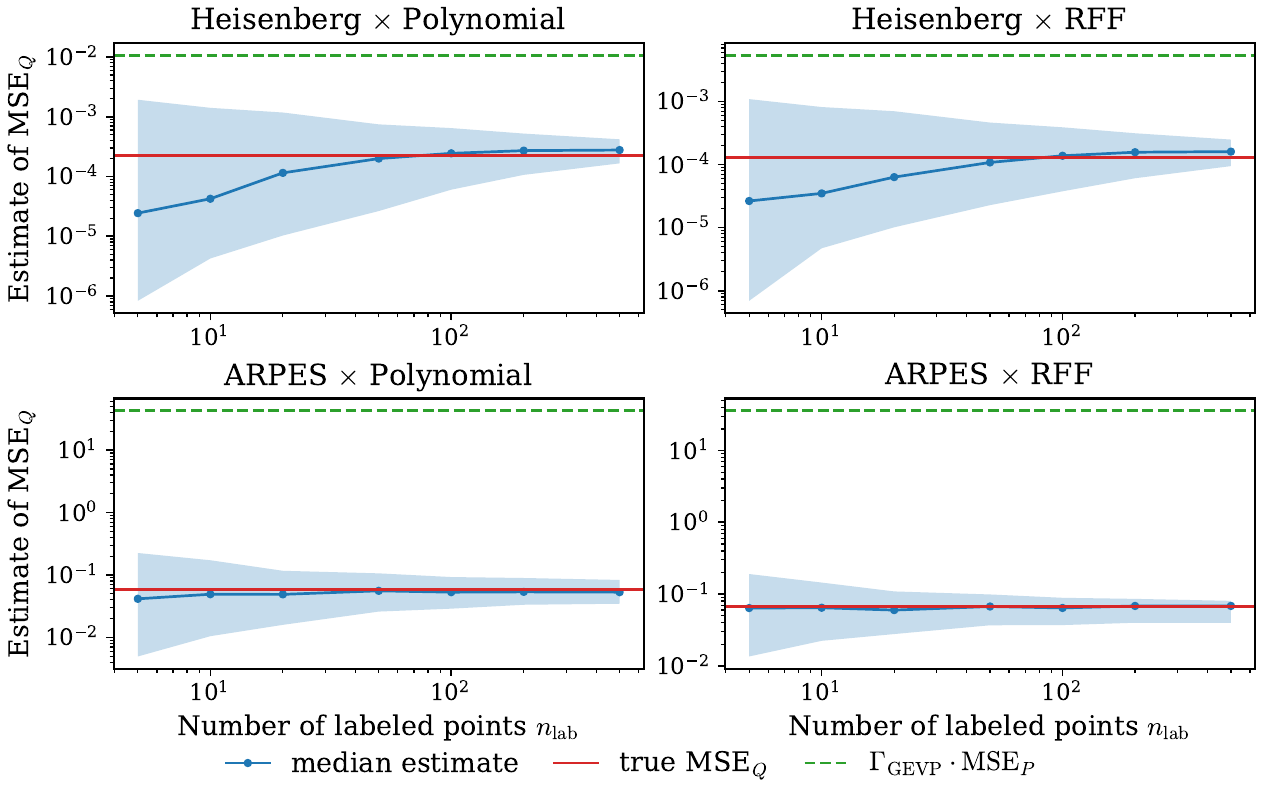}
    \caption{Direct estimation of $\mathrm{MSE}_Q$ from a labeled budget $n_{\mathrm{lab}}$
    drawn uniformly from $Q$. Median empirical estimate and $5$--$95$ percentile range over
    $200$ draws, against the true $\mathrm{MSE}_Q$ and
    $\Gamma_{\mathrm{GEVP}} \cdot \mathrm{MSE}_P$. }
    \label{fig:labeled-budget}
\end{figure}
The alternative is to measure a few points inside the target phase and estimate $\mathrm{MSE}_Q$ directly. \Cref{fig:labeled-budget} shows why this fails at small budgets. A few points drawn uniformly usually underestimate the error. On Heisenberg, twenty labeled points still fall below half the true error more often than not. The bias disappears only at several hundred points. 

Both alternatives underestimate the error. The posterior variance falls further behind as the extrapolation deepens. The labeled estimate stays biased downward until several hundred points are measured. The GEVP bound is conservative, but it never falls below the measured error.

\section{Datasets}
\label{app:datasets}

\paragraph{Heisenberg.} We use the mean-field Landau--Ginzburg free energy $f(T,m)=(1-T_c/T)\,m^2/2+(T_c/T)^3\,m^4/12+c_0,$ with $T_c=1.5$ and an additive offset $c_0=0.01$. Here $c_0$ is an overall constant shift to the free energy and therefore has no effect on the location of its minima. The training and extrapolation rectangles are $P = (2.0, 3.0) \times (-1.25, 1.25)$ and $Q = (1.0, 2.0) \times (-1.25, 1.25)$, so $\rho_P = \rho_Q = 2$.

\vspace{-3mm}

\paragraph{Flory--Huggins.} We use the mixing free energy $f(T,x)=T[x\ln x+(1-x)\ln(1-x)]+2T_cx(1-x),$ with $T_c=1$, temperature $T$, and volume fraction $x\in(0,1)$. The critical point occurs at equal composition $x=1/2$, so we use $m=x-1/2$ as the coordinate. Below $T_c$, the coexisting phases lie at $\pm m^*(T)$, and the order parameter is $|m^*(T)|=\left|\arg\min_m f(T,m+1/2)\right|$. The rectangles are $P = (1.5, 2.5) \times (-0.48, 0.48)$ and $Q = (0.5, 1.5) \times (-0.48, 0.48)$
in $(T, m)$, so $\rho_P = \rho_Q = 2$.

\vspace{-3mm}

\paragraph{SSH polaron.} We take the Hamiltonian and coupling convention of \citet{marchand2010sharp}, the source of the benchmark in \citet{vargas2018extrapolating}: hopping $t = 1$, dispersionless phonons of frequency $\Omega$, and $\lambda_{\mathrm{SSH}} = 2\alpha^2/(t\Omega)$, where $\alpha$ is the electron--phonon coupling strength. We use $\Omega = 3.0$, one of the two frequencies they study.  We compute the dispersion numerically using variational exact diagonalization in the total-momentum sector, with the phonon cloud confined to within four sites of the electron and at most ten phonons. Ground truth is a grid of $43$ couplings $\times$ $61$ momenta, interpolated with a cubic spline. The critical coupling on this grid is $\lambda_c = 0.6839$. The rectangles are $P = (0.2, 0.6) \times (0, 0.6\pi)$ and $Q = (0.6, 1.0) \times (0, 0.6\pi)$, so $\rho_P = \rho_Q = 2$.

\vspace{-3mm}

\paragraph{Rayleigh--B\'enard.} The mean temperature profile $\bar T(\mathrm{Ra}, z)$ comes from a sweep of numerical simulations of two-dimensional convection over $\mathrm{Ra} \in [300, 3 \times 10^4]$. We use $\log_{10}\mathrm{Ra}$ as the tuning parameter, which keeps the polynomial and random-feature maps well conditioned. Ground truth is a grid of $64$ Rayleigh numbers $\times$ $49$ heights, interpolated with a cubic spline. The rectangles are $P = (\log_{10} 1500, \log_{10} 7500) \times (0, 1)$ and $Q = (\log_{10} 300, \log_{10} 1500) \times (0, 1)$, of equal log-width, so $\rho_P = \rho_Q = 2$. The first convecting run is at $\mathrm{Ra} = 835$, and a linear fit of $\mathrm{Nu} - 1$ against $\mathrm{Ra}$, where $\mathrm{Nu}$ is the Nusselt number, over the first convecting runs gives an onset of $\mathrm{Ra} \approx 776.2$. We therefore set $T_c = \log_{10} 776.2$ instead of using the free-slip linear-theory value $27\pi^4/4 \approx 657.5$~\citep{chandrasekhar1961hydrodynamic}, which the simulation does not realize.

\vspace{-3mm}

\paragraph{ARPES.} We use antinodal energy-distribution curves (EDCs) of overdoped Pb-doped Bi-2212 from \citet{he2021rapid}, taking the central of the three momentum-integration windows provided. The raw $(T, E)$ grid has $100$ temperatures non-uniformly spaced from $13.5$ K to $290$ K and $511$ binding energies $E_j = -0.371173 + 0.001\,j$ eV, $j = 0, \dots, 510$. Ground truth is a linear interpolation on this grid, with NaN values outside the analysis window replaced by column medians. The rectangles are $P = (95, 150)\,\mathrm{K} \times (-0.1, 0.0)\,\mathrm{eV}$ and $Q = (65, 95)\,\mathrm{K} \times (-0.1, 0.0)\,\mathrm{eV}$, so $\rho_P \approx 1.55$ and $\rho_Q \approx 2.83$. $Q$ spans the fluctuation regime between $T_c \approx 66$ K and $T_{\mathrm{gap}} \approx 90$ K.

\smallskip

\noindent \textit{ARPES preprocessing sensitivity.} The default pipeline applies no smoothing or denoising. The target is the raw Fermi-function-divided intensity, linearly interpolated on the measured grid, and peak energies are extracted by a Gaussian fit. To check that the recovered peak shift is not an artifact of these choices, we vary one setting at a time and recompute both the fit and $E^*(T)$ (\Cref{tab:arpes-sensitivity}). Changes to interpolation and smoothing shift the recovered $E^*(T)$ by at most $0.11$ meV, compared with a $4.8$ meV shift under the default pipeline. The peak-finding rule matters more. Replacing the Gaussian fit with a grid argmax or a parabolic refinement reduces the peak shift by about $15\%$, because both are biased by the $1$ meV energy sampling. We therefore use the Gaussian fit.

\begin{table}[!t]
    \centering
    \small
    \caption{ARPES preprocessing sensitivity at $n = 2000$ with fixed hyperparameters, averaged over 5 seeds. $\max|\Delta E^*|$ is the largest deviation of the peak-energy curve from the default, in meV. $\Delta$shift is the change in peak-shift magnitude.}
    \label{tab:arpes-sensitivity}
    \vspace{0.4em}
    \setlength{\tabcolsep}{4pt}
    \footnotesize
    \begin{tabular}{lccccc}
        \toprule
        & & & \multicolumn{3}{c}{Extrap NMSE} \\
        \cmidrule(lr){4-6}
        Variant & $\max|\Delta E^*|$ & $\Delta$shift & Poly & RFF & GP \\
        \midrule
        default                    & $0.00$ & $\phantom{-0}0\%$ & $0.0382$ & $0.0417$ & $0.0082$ \\
        \midrule
        cubic interpolation        & $0.03$ & $-0.2\%$          & $0.0439$ & $0.0477$ & $0.0083$ \\
        energy window $-0.08$\,eV  & $0.00$ & $\phantom{-0}0\%$ & $0.0638$ & $0.0450$ & $0.0098$ \\
        energy window $-0.12$\,eV  & $0.00$ & $\phantom{-0}0\%$ & $0.0270$ & $0.0441$ & $0.0072$ \\
        Savitzky--Golay smoothing  & $0.11$ & $+0.3\%$          & $0.0365$ & $0.0434$ & $0.0072$ \\
        peak: grid argmax          & $2.53$ & $-15.3\%$         & $0.0382$ & $0.0417$ & $0.0082$ \\
        peak: parabolic            & $2.50$ & $-15.2\%$         & $0.0382$ & $0.0417$ & $0.0082$ \\
        \bottomrule
    \end{tabular}
\end{table}

\vspace{-3mm}

\paragraph{Sampling.} On Heisenberg, Flory--Huggins, SSH, and Rayleigh--B\'enard, training points are drawn independently and uniformly on $P$ for each seed. For ARPES, we stratify in $T$ across $\sqrt{n}$ equal-width strata and draw uniformly in $E$, which keeps the non-uniformly spaced temperature axis covered at small $n$. For every target, a fixed set of $2000$ points drawn uniformly on $Q$ with seed $99$ is held out for evaluation and shared across all methods, so that any difference in extrapolation error comes from the model rather than the test sample.

\vspace{-3mm}

\paragraph{Training-size sweep (\Cref{fig:bounds-vs-n}).} For each method, target, and seed, we train on ten linearly spaced sizes $n \in \{200, 400, \dots, 2000\}$. Hyperparameters are selected on the training split at $n = 2000$ for each of the five seeds, and the most frequently chosen configuration is used at every $(n, \text{seed})$. This isolates the effect of training size.

\vspace{-3mm}

\paragraph{Distance sweep (\Cref{fig:power-law}).} We move the training rectangle away from the transition so that its near edge lies at distance $D$ from $T_c$, keeping its width fixed so that the training density at fixed $n$ does not change with $D$. The extrapolation rectangle $Q$ stays at its default position. Each target uses a ten-point geometric grid in $D$ at $n = 2000$ with five seeds per point, spanning $D \in [0.5, 3.0]$ on Heisenberg and Flory--Huggins, $[5, 130]$ K on ARPES, $[0.1, 0.484]$ in $\lambda_{\mathrm{SSH}}$ on SSH, and $[0.286, 0.888]$ in $\log_{10}\mathrm{Ra}$ on Rayleigh--B\'enard. The grid is geometric so that its points are evenly spaced on the logarithmic axis used to fit the power laws.

\section{Neural Network Methods}
\label{app:nn}

The MLP failure in \Cref{sec:methods} is a property of the architecture. Past its outermost kink, a ReLU network extrapolates along a line whose slope is not fixed by the training data. We benchmark four architectures with more structure built in, two that change the representation and two that build in the physics. Implementation details are in \Cref{app:training}.

\smallskip
\noindent \textbf{Fourier-feature MLP}~\citep{tancik2020fourier}. The input is embedded in sinusoids at fixed random frequencies before it reaches the network. A plain MLP is biased toward low frequencies~\citep{rahaman2019spectral}. The embedding counteracts this bias. Because the embedding is periodic, the output does not become affine beyond the data.

\smallskip
\noindent \textbf{SIREN}~\citep{sitzmann2020siren}. Every activation is a sinusoid $\sin(\omega_0\,\cdot)$, so the network is a composition of periodic functions. It removes affine extrapolation through the  architecture rather than the input encoding. Beyond the data, SIREN therefore oscillates instead of extrapolating linearly, at a frequency set by $\omega_0$ and the initialization.

\smallskip
\noindent \textbf{PINN}~\citep{raissi2019physics}. The architecture is the vanilla MLP, but the loss adds penalties for violating known physics on unlabeled collocation points spanning $P$ and $Q$. For Heisenberg the penalty enforces the $\mathbb{Z}_2$ parity $f(T, m) = f(T, -m)$. For ARPES it enforces positivity of the intensity and approximate conservation of the energy-integrated spectral weight. The constraints require no labels, so enforcing them on $Q$ uses no data from the target phase.

\smallskip
\noindent \textbf{Gray-box}~\citep{psichogios1992hybrid}. The physical functional form is hard-coded. A small network learns only its coefficients, as smooth functions of the tuning parameter. For Heisenberg the form is the Landau expansion $f_\theta(T, m) = a_\theta(T)\, m^2 + b_\theta(T)\, m^4 + c_\theta(T)$. For ARPES it is a single spectral peak on a linear background, with amplitude, center, and width predicted by the network as functions of $T$. The network then extrapolates only the coefficient functions, not the observable itself.

\begin{table}[!ht]
    \centering
    \small
    \caption{Neural network results on the same protocol as \Cref{tab:metrics}, averaged over five seeds. A dash means no critical point was recovered.}
    \label{tab:metrics-neural}
    \vspace{0.4em}
    \begin{tabular}{lccc|cc}
        \toprule
        & \multicolumn{3}{c}{\textbf{Heisenberg}} & \multicolumn{2}{c}{\textbf{ARPES}} \\
        \cmidrule(lr){2-4} \cmidrule(lr){5-6}
        Method & Extrap NMSE & Extrap Rel.\ \% & $|\Delta T_c|/T_c$ (\%) & Extrap NMSE & Extrap Rel.\ \% \\
        \midrule
        Vanilla MLP & $2.0\!\times\!10^{-1}$ & $\phantom{0}28.8$ & $15.7$  & $1.1\!\times\!10^{-1}$ & $23.1$ \\
        FF-MLP      & $4.2\!\times\!10^{0}$ & $117.6$           & $10.5$  & $9.4\!\times\!10^{-2}$ & $19.6$ \\
        SIREN       & $5.3\!\times\!10^{0}$ & $168.0$           & $23.0$  & $1.1\!\times\!10^{0}$  & $79.0$ \\
        PINN        & $1.8\!\times\!10^{0}$ & $116.7$           & ---     & $1.9\!\times\!10^{-1}$ & $28.3$ \\
        Gray-box    & $1.2\!\times\!10^{-1}$ & $\phantom{0}22.1$ & ---     & $1.0\!\times\!10^{1}$ & $141.4$ \\
        \bottomrule
    \end{tabular}
\end{table}
\FloatBarrier

\begin{figure}[!ht]
    \centering
    \includegraphics[width=0.8\linewidth]{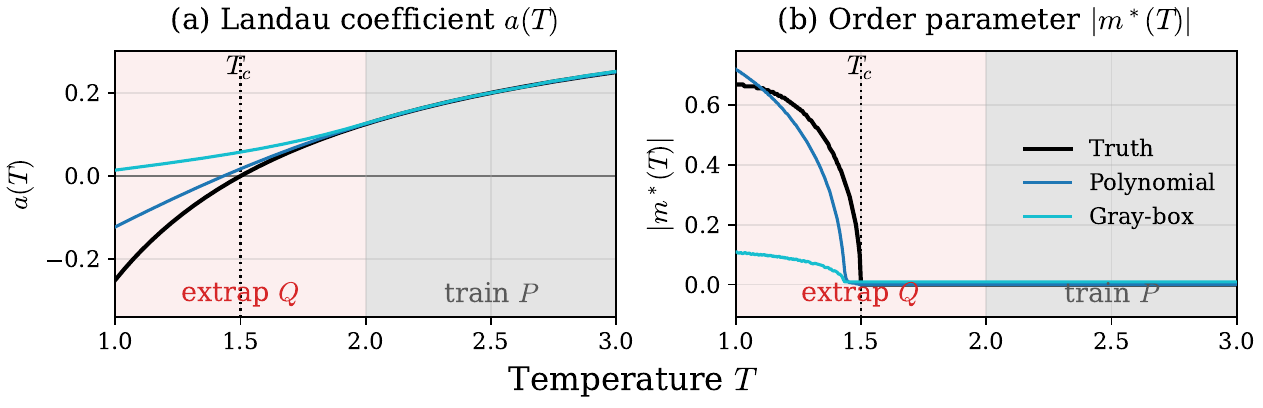}
    \caption{Landau coefficient $a(T) = \tfrac{1}{2}\partial_m^2 f|_{m=0}$ (a) and the resulting order parameter $|m^*(T)|$ (b) of the fitted Heisenberg surfaces, averaged over five seeds. The double well requires $a < 0$.}
    \label{fig:landau-coeff}
\end{figure}

\begin{figure}[!ht]
    \centering
    \includegraphics[width=\linewidth]{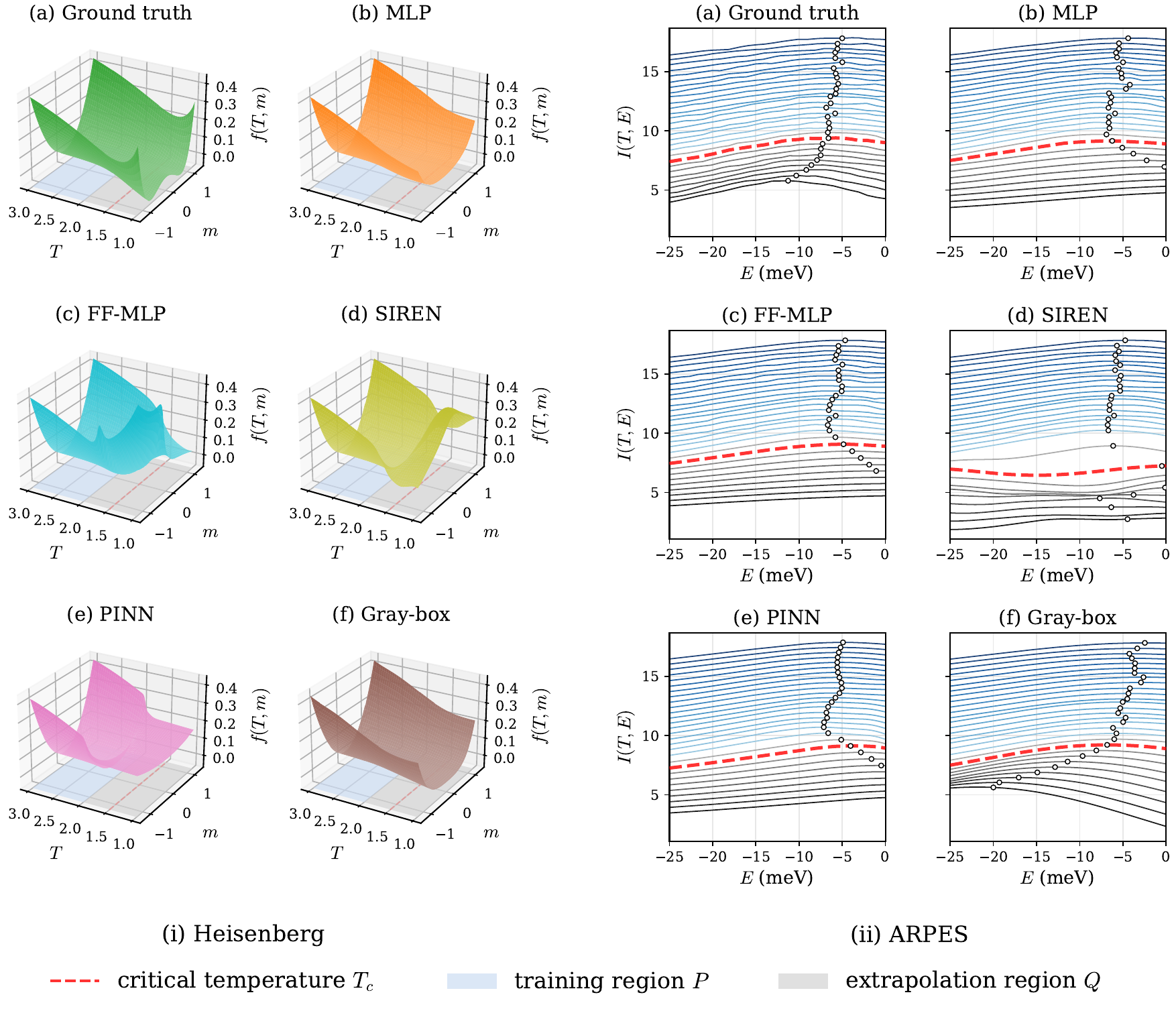}
    \caption{Surfaces learned by the neural-network models on Heisenberg (free energy $f(T, m)$) and ARPES (intensity $I(T, E)$), compared with the ground truth.}
    \label{fig:landscape-neural}
\end{figure}

\vspace{-3mm}
\paragraph{Results.} None of the four architectures recovers the unobserved phase (\Cref{tab:metrics-neural}). On Heisenberg, the Fourier-feature MLP and SIREN do worse than the vanilla MLP. The Fourier-feature MLP introduces high-frequency artifacts below $T_c$, and SIREN produces a strongly oscillating surface with no resemblance to the double well (\Cref{fig:landscape-neural}). The PINN's surface is even in $m$, so it satisfies its parity constraint, but it still does not develop the double well. The gray-box is the strongest of the four. It improves the extrapolation over the vanilla MLP, yet it too fails to develop the double well. A quartic in $m$ has a double well exactly when its $m^2$ coefficient is negative, so the well requires $a_\theta(T)$ to change sign at $T_c$, and $a_\theta$ is a small MLP in $T$ trained only where it is positive. Beyond the data it behaves like any MLP and never crosses zero (\Cref{fig:landau-coeff}). The polynomial has no separate coefficient to extrapolate. Its $T$-dependence is part of one fixed basis, so the fit inside $P$ determines the surface everywhere, and the fitted trend carries the $m^2$ coefficient through zero.

On ARPES no architecture recovers the peak shift (\Cref{fig:landscape-neural}). The Fourier-feature MLP and the PINN are within a factor of two of the vanilla MLP, and SIREN is an order of magnitude worse. The gray-box, whose spectral-peak form is the most physical of the four, is the least stable across seeds and has the largest error. Across both targets, every neural variant trails polynomial regression by a factor of two to twenty in relative error, and those that locate a critical point at all miss it by two to five times the polynomial's margin. Each architecture replaces the MLP's affine extrapolation with a different bias beyond the data, and none recovers the target. This does not completely rule out a network that extrapolates here. It shows that these architectures, though designed with extrapolation in mind, fail on these targets, and that their design gives no indication of when they would succeed.

\section{Training Details}
\label{app:training}

Every hyperparameter is chosen on the training region alone. Each run draws $n = 2000$ points from $P$, holds out $20\%$, and fits on the rest. Polynomial, RFF, and GP hyperparameters are selected by maximizing the log marginal likelihood (type-II maximum likelihood). Neural networks are selected by lowest holdout MSE. Every result is a mean over five seeds $\{42, \dots, 46\}$, each a fresh draw of the $n$ points. The grids below are shared across targets and narrowed for a few. Selected settings appear in \Cref{tab:hp-selected} and \Cref{tab:hp-selected-neural}.

\vspace{-3mm}

\paragraph{Polynomial.} Total degree $d \in \{2, 4, 6, 8, 10, 12\}$ and ridge weight $\alpha \in \{10^{-8}, 10^{-6}, 10^{-4}, 10^{-2}\}$. Landau theory motivates the choice of class as a low-order expansion about the critical point, so on three targets we fix the degree on physical grounds instead of searching. On SSH the degree is fixed at $4$. The dispersion is even in $K$ and its minimum leaves the origin continuously~\citep{marchand2010sharp}, and a quartic is the lowest even order that can do this. On Rayleigh--B\'enard the degree in $\log_{10}\mathrm{Ra}$ is fixed at $2$ because the profile varies smoothly along it. Only the $z$ degree is searched. On ARPES the degree is fixed at $10$. Lower degrees miss the spectral peak inside $P$, so they cannot be expected to extrapolate.

\vspace{-3mm}

\paragraph{RFF.} $m \in \{1000, 2000, 4000, 8000\}$ random features, per-axis length scales equal to the median pairwise distance times a factor in $\{0.25, 0.5, 1, 2, 4\}$, and ridge $\alpha \in \{10^{-8}, 10^{-6}, 10^{-4}, 10^{-2}\}$. The log marginal likelihood is averaged over three feature draws. On targets where the median underestimates the length scale along the tuning axis, the factor grid for that axis extends upward.

\vspace{-3mm}

\paragraph{GP.} \texttt{scikit-learn}'s \texttt{GaussianProcessRegressor} with \texttt{normalize\_y=True} and noise $10^{-8}$. Base kernels $\{\mathrm{LIN}, \mathrm{RBF}, \mathrm{MAT5/2}, \mathrm{RQ}\}$ with anisotropic length scales, combined by a depth-wise greedy search to depth at most $3$. Each candidate is fit by maximizing the log marginal likelihood with one L-BFGS run and scored by BIC~\citep{schwarz1978estimating}, and the selected kernel is refit with three restarts.

\vspace{-3mm}

\paragraph{Neural networks.} All five architectures share one protocol. Training uses AdamW, reduces the learning rate tenfold whenever the holdout MSE has not improved for $500$ epochs, and stops when the learning rate falls below $10^{-7}$ or after $5 \times 10^{5}$ epochs. We retain the checkpoint with lowest holdout MSE and fix the selected configuration across five seeds.

\smallskip

\noindent\emph{MLP.} Fully connected with skip connections~\citep{he2016deep} and LayerNorm~\citep{ba2016layernormalization}. Depth $\in \{2, 4, 6, 8\}$, width $\in \{8, 128, 256, 512\}$, activation $\in \{\text{GELU}, \tanh\}$, learning rate $\in \{10^{-2}, 3 \times 10^{-3}\}$, weight decay $\in \{10^{-5}, 10^{-4}\}$. ReLU was excluded after a preliminary sweep.

\smallskip

\noindent\emph{FF-MLP.} We encode inputs as $\gamma(x) = (\cos 2\pi Bx,\ \sin 2\pi Bx)$, with $B \in \mathbb{R}^{m_{\mathrm{ff}} \times 2}$ sampled once from $\mathcal{N}(0, \sigma^2)$~\citep{tancik2020fourier}. The MLP grid, plus $m_{\mathrm{ff}} \in \{64, 128, 256\}$ and $\sigma \in \{0.05, 0.1, 0.5, 1.0\}$.

\smallskip

\noindent\emph{SIREN.} Every activation is $\sin(\omega_0\,\cdot)$~\citep{sitzmann2020siren}. The MLP depth and width grid with first-layer $\omega_0 \in \{2, 5, 10\}$, hidden-layer $\omega_0 \in \{1, 5, 10\}$, learning rate $\in \{10^{-4}, 10^{-3}\}$, and no weight decay.

\smallskip

\noindent\emph{PINN.} The vanilla MLP fixed at $[128{\times}4]$ GELU, learning rate $\eta = 3 \times 10^{-3}$, weight decay $10^{-4}$, trained on $\mathrm{MSE}_P + \lambda_{\mathrm{phys}}\,\mathcal{R}$ with $\lambda_{\mathrm{phys}} \in \{1, 10\}$. On Heisenberg $\mathcal{R} = \langle (f(T,m) - f(T,-m))^2 \rangle$ enforces $\mathbb{Z}_2$ parity. On ARPES $\mathcal{R} = \langle \max(0, -f)^2 \rangle + \mathrm{Var}_T\!\left[\langle f \rangle_E\right]$ enforces positivity of the intensity and a temperature-independent energy-integrated spectral weight. The averages run over $4096$ collocation points drawn uniformly on $P \cup Q$, and the integrated weight over a $64 \times 64$ grid.

\smallskip

\noindent\emph{Gray-box.} On Heisenberg, we use a quartic Landau expansion, $f_\theta(T, m) = a_\theta(T)\,m^2 + b_\theta(T)\,m^4 + c_\theta(T)$. On ARPES, we use a Gaussian peak on a linear background, $f_\theta(T, E) = A_\theta(T)\exp\!\left(-\tfrac{(E - E_{0,\theta}(T))^2}{2 w_\theta(T)^2}\right) + \beta_{0,\theta}(T) + \beta_{1,\theta}(T)\,E.$ A softplus keeps $A$ and $w$ positive, and a sigmoid keeps $E_0$ inside the measured window. Each coefficient function is a network of depth $2$ and width $64$, with $\eta \in \{10^{-3}, 3 \times 10^{-3}\}$ and weight decay $10^{-4}$.

\vspace{-3mm}

\paragraph{Compute.} Polynomial, RFF, and GP training and the bound computations ran on CPU partitions (4--16 cores, 16--64\,GB, minutes per sweep cell). Neural network training ran on H200 GPUs at roughly ten minutes per seed. Including hyperparameter selection and both sweeps, the experiments used a few hundred CPU-hours and a few hundred GPU-hours.
\begin{table}[!h]
    \centering
    \small
    \caption{Selected hyperparameters for the four methods across all targets. RFF length scales are in standardized units, ordered as (tuning axis, second axis). Rayleigh--B\'enard uses per-axis polynomial degree caps $(d_{\mathrm{tuning}}, d_z)$ in $\log_{10}\mathrm{Ra}$ and $z$. $[w{\times}L]$ denotes width and depth, and $\eta$ the learning rate.}
    \label{tab:hp-selected}
    \setlength{\tabcolsep}{3pt}
    \footnotesize
    \begin{tabular}{llllll}
        \toprule
        & Heisenberg & ARPES & Polaron & Flory--Huggins & Rayleigh--B\'enard \\
        \midrule
        Poly. & $d = 8$ & $d = 10$ & $d = 4$ & $d = 8$ & $(2, 12)$ \\
              & $\alpha = 10^{-6}$ & $\alpha = 10^{-4}$ & $\alpha = 10^{-2}$ & $\alpha = 10^{-4}$ & $\alpha = 10^{-8}$ \\
        \addlinespace
        RFF & $m = 8000$ & $m = 2000$ & $m = 4000$ & $m = 8000$ & $m = 8000$ \\
            & $\ell = (3.6, 0.45)$ & $\ell = (3.6, 0.44)$ & $\ell = (57, 0.89)$ & $\ell = (7.2, 0.45)$ & $\ell = (1.8, 0.45)$ \\
            & $\alpha = 10^{-8}$ & $\alpha = 10^{-4}$ & $\alpha = 10^{-8}$ & $\alpha = 10^{-8}$ & $\alpha = 10^{-8}$ \\
        \addlinespace
        GP & $\mathrm{RBF}{\cdot}\mathrm{LIN}^{3}$ & $\mathrm{MAT5/2} + \mathrm{RBF}$ & $\mathrm{RBF} + \mathrm{MAT5/2}$ & $\mathrm{RBF}$ & $\mathrm{MAT5/2} \cdot \mathrm{MAT5/2}$ \\
        \addlinespace
        MLP & $[256{\times}2]$, GELU & $[512{\times}4]$, GELU & $[128{\times}4]$, GELU & $[128{\times}8]$, $\tanh$ & $[128{\times}8]$, $\tanh$ \\
            & $\eta = 3 \times 10^{-3}$ & $\eta = 3 \times 10^{-3}$ & $\eta = 10^{-2}$ & $\eta = 10^{-2}$ & $\eta = 10^{-2}$ \\
        \bottomrule
    \end{tabular}
\end{table}

\begin{table}[!htbp]
    \centering
    \small
    \caption{Selected hyperparameters for the additional neural baselines, run on Heisenberg and ARPES only (\Cref{app:nn}). $[w{\times}L]$ denotes width $w$ and depth $L$, $\eta$ the learning rate, and wd the weight decay.}
    \label{tab:hp-selected-neural}
    \setlength{\tabcolsep}{5pt}
    \footnotesize
    \begin{tabular}{lll}
        \toprule
        & Heisenberg & ARPES \\
        \midrule
        FF-MLP & $m_{\mathrm{ff}} = 128$, $\sigma = 0.1$, $[256{\times}4]$, GELU
               & $m_{\mathrm{ff}} = 256$, $\sigma = 0.05$, $[256{\times}6]$, GELU \\
               & $\eta = 3 \times 10^{-3}$, wd $= 10^{-4}$
               & $\eta = 10^{-3}$, wd $= 10^{-4}$ \\
        \addlinespace
        SIREN & $[256{\times}4]$, $\omega_0 = (2,\,1)$
              & $[256{\times}4]$, $\omega_0 = (10,\,5)$ \\
              & $\eta = 10^{-4}$, wd $= 0$
              & $\eta = 10^{-4}$, wd $= 0$ \\
        \addlinespace
        PINN & $[128{\times}4]$, GELU, $\lambda_{\mathrm{phys}} = 1$
             & $[128{\times}4]$, GELU, $\lambda_{\mathrm{phys}} = 1$ \\
             & $\mathbb{Z}_2$ parity; $\eta = 3 \times 10^{-3}$, wd $= 10^{-4}$
             & positivity $+$ sum rule; $\eta = 3 \times 10^{-3}$, wd $= 10^{-4}$ \\
        \addlinespace
        Gray-box & Landau $m^2, m^4$; $[64{\times}2]$
                 & Gaussian peak $+$ linear bg.; $[64{\times}2]$ \\
                 & $\eta = 3 \times 10^{-3}$, wd $= 10^{-4}$
                 & $\eta = 10^{-3}$, wd $= 10^{-4}$ \\
        \bottomrule
    \end{tabular}
\end{table}

\clearpage
\section{Supplementary Figures}
\label{app:figures}
\begin{figure}[!h]
    \centering
    \includegraphics[width=\linewidth,height=0.78\textheight,keepaspectratio]{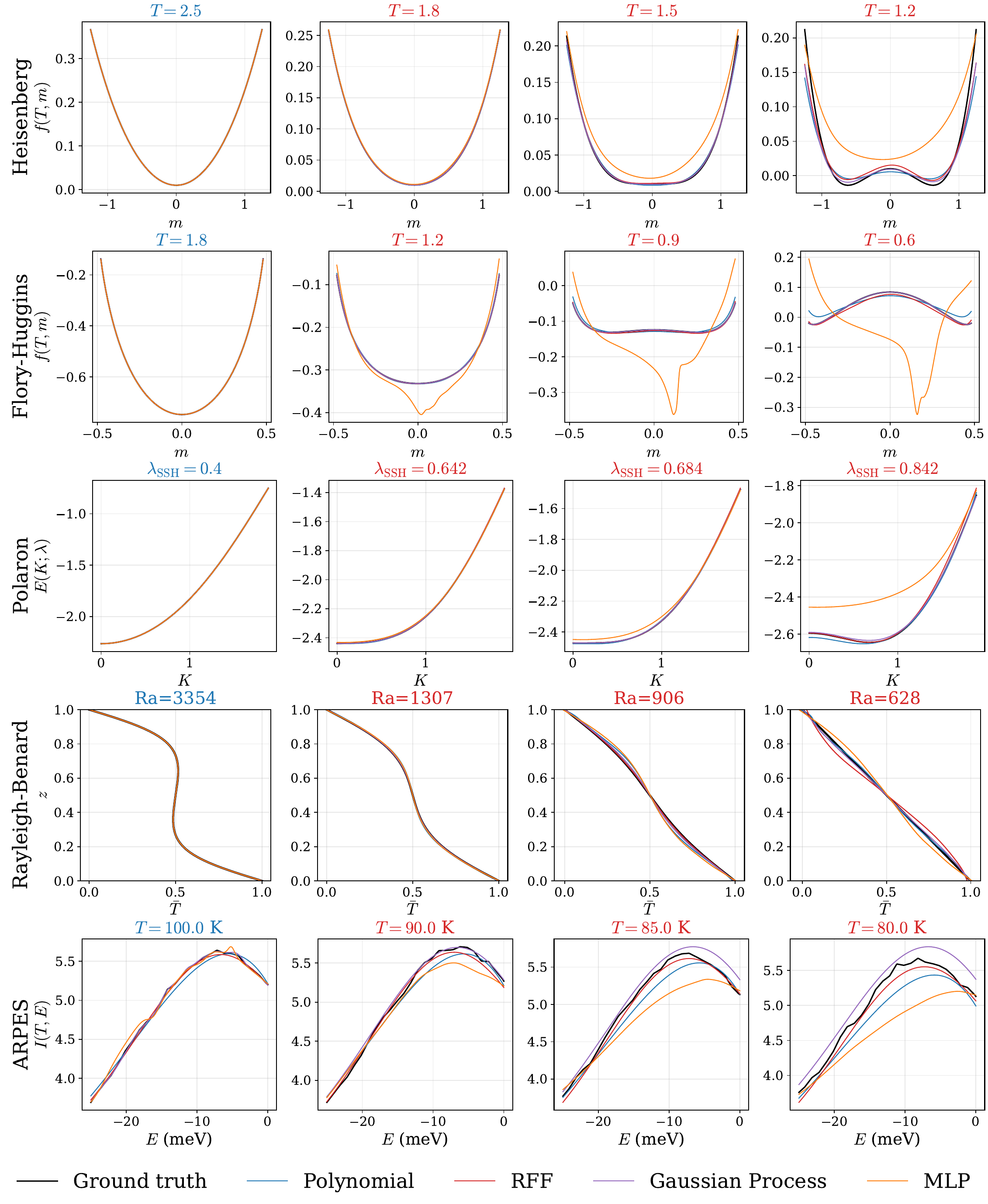}
    \caption{Slices at four fixed values of the tuning parameter for each of the five targets, one row per target. One slice lies inside the training region and three span the transition. Curves show the ground truth and the predictions of each method. Generated by slicing the predicted surfaces of \Cref{fig:landscape}.}
\label{fig:slices}
\end{figure}

\end{document}